\documentclass[10pt,journal,compsoc]{IEEEtran}
\ifCLASSOPTIONcompsoc
  \usepackage[nocompress]{cite}
\else
  \usepackage{cite}
\fi
\usepackage{amsmath,amsfonts}
\usepackage{algorithmic}
\usepackage{algorithm}
\usepackage{array}
\usepackage[caption=false,font=normalsize,labelfont=sf,textfont=sf]{subfig}
\usepackage{textcomp}
\usepackage{stfloats}
\usepackage{url}
\usepackage{verbatim}
\usepackage{graphicx}
\usepackage{cite}
\usepackage{boldline,multirow}
\usepackage{booktabs}
\usepackage{diagbox}
\usepackage{xcolor}
\begin{document}
\newcommand{\lsx}[1]{{\bf\color{cyan}[{\sc Lsx:} #1]}}
\newcommand{\CP}[1]{{\bf\color{red}[{\sc CP:} #1]}}
\newcommand{\FC}[1]{{\bf\color{red}[{\sc FC:} #1]}}
\newcommand{\KC}[1]{{\bf\color{red}[{\sc KC:} #1]}}
\newcommand{\YS}[1]{{\bf\color{red}[{\sc YS:} #1]}}
\newcommand{\ZBX}[1]{{\bf\color{red}[{\sc ZBX:} #1]}}

\newcommand{\model}{SchemaWalk}

\title{Inductive Meta-path Learning for Schema-complex Heterogeneous Information Networks}

\author{Shixuan Liu*, Changjun Fan*, Kewei Cheng*, Yunfei Wang, Peng Cui,~\IEEEmembership{Senior Member,~IEEE},\\ Yizhou Sun,~\IEEEmembership{Member,~IEEE}, Zhong Liu
        % <-this % stops a space

\IEEEcompsocitemizethanks{
\IEEEcompsocthanksitem Shixuan Liu, Changjun Fan and Zhong Liu are with the Laboratory for Big Data and Decision, College of Systems Engineering, National University of Defense Technology, Hunan, China. E-mail: \{liushixuan, fanchangjun, liuzhong\}@nudt.edu.cn. 
\IEEEcompsocthanksitem Yunfei Wang is with the National Key Laboratory of Information Systems Engineering, College of Systems Engineering, National University of Defense Technology, Hunan, China. E-mail: wangyunfei@nudt.edu.cn.
\IEEEcompsocthanksitem Kewei Cheng and Yizhou Sun are with the University of California, Los Angeles, United States. E-mail: \{viviancheng, yzsun\}@cs.ucla.edu.% <-this % stops a space
\IEEEcompsocthanksitem Peng Cui is with the Department of Computer Science and Technology, Tsinghua University, Beijing, China. E-mail: cuip@tsinghua.edu.cn.% <-this % stops a space
\IEEEcompsocthanksitem 
Our Code is available at https://github.com/shixuanliu-andy/SchemaWalk}
\thanks{*These authors contributed equally.}% <-this % stops a space
\thanks{(Corresponding authors: Changjun Fan, Kewei Cheng, Zhong Liu.)}% <-this % stops a space
}

\markboth{IEEE TRANSACTIONS ON PATTERN ANALYSIS AND MACHINE INTELLIGENCE}%
{Liu \MakeLowercase{\textit{et al.}}: Inductive Meta-path Learning for Schema-complex Heterogeneous Information Networks}


\IEEEtitleabstractindextext{
\begin{abstract}

Heterogeneous Information Networks (HINs) are information networks with multiple types of nodes and edges. The concept of meta-path, i.e., a sequence of entity types and relation types connecting two entities, is proposed to provide the meta-level explainable semantics for various HIN tasks. Traditionally, meta-paths are primarily used for schema-simple HINs, e.g., bibliographic networks with only a few entity types, where meta-paths are often enumerated with domain knowledge.
However, the adoption of meta-paths for schema-complex HINs, such as knowledge bases (KBs) with hundreds of entity and relation types, has been limited due to the computational complexity associated with meta-path enumeration. Additionally, effectively assessing meta-paths requires enumerating relevant path instances, which adds further complexity to the meta-path learning process.
% However, only few efforts have been made on adopting meta-paths for schema-complex HINs, e.g., knowledge bases (KBs) with hundreds of entity and relation types, since meta-path enumeration in such HINs is computationally-prohibitive.
% Furthermore, assessing meta-paths effectively requires enumerating the relevant path instances, which also adds complexity to meta-path learning. 
To address these challenges, we propose~\model, an inductive meta-path learning framework for schema-complex HINs.
We represent meta-paths with schema-level representations to support the learning of the scores of meta-paths for varying relations, mitigating the need of exhaustive path instance enumeration for each relation.
Further, we design a reinforcement-learning based path-finding agent, which directly navigates the network schema (i.e., schema graph) to learn policies for establishing meta-paths with high coverage and confidence for multiple relations.
% By representing meta-paths with schema-level representations, the scores of meta-paths for varying relations could be learned, mitigating the need to enumerate path instance for every relation during learning.
Extensive experiments on real data sets demonstrate the effectiveness of our proposed paradigm. 
\end{abstract}

}
\maketitle
\section{Introduction}
\label{intro}
\begin{figure*}[h!]
\centering
\includegraphics[width=0.8\linewidth]{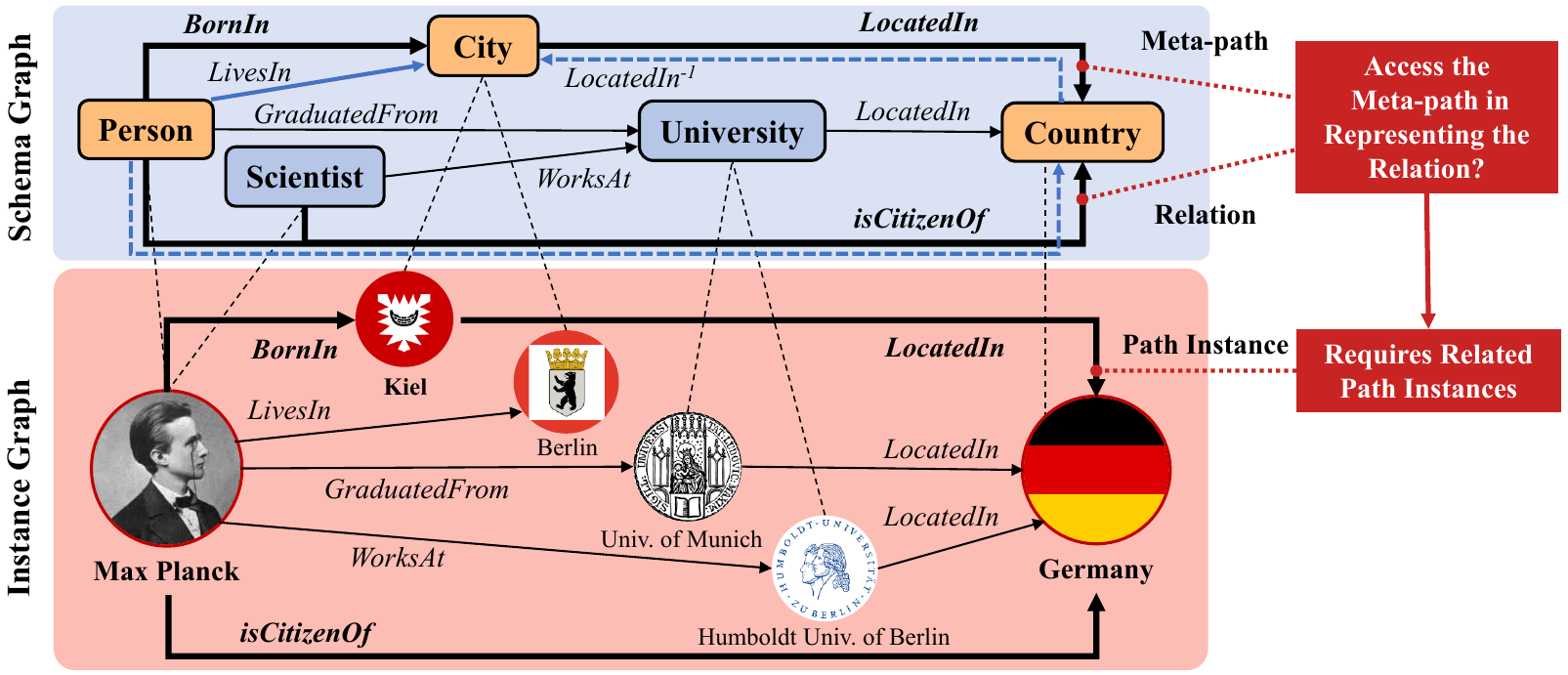}
\caption{An illustration of a two-view heterogeneous information network. \textbf{Top:} a schema graph $\mathcal{G}_S$ where nodes represent entity types. \textbf{Bottom:} an instance graph $\mathcal{G}_I$ where nodes represent specific entities. The entities in $\mathcal{G}_I$ are linked to one or multiple entity types in $\mathcal{G}_S$. 
A path in the $\mathcal{G}_S$ (one of which is boldfaced in black) is referred to as a meta-path.
To learn these meta-paths, evidence from corresponding path instances, marked in black bold line in $\mathcal{G}_I$, is typically necessary.
}\label{fig1}
\end{figure*}

\IEEEPARstart{H}{eterogeneous} information networks (HINs), such as DBPedia~\cite{auer2007dbpedia}, amazon product graph~\cite{dong2018challenges} and protein data bank~\cite{berman2000protein}, have grown rapidly recently and provide remarkably valuable resources for many real-world applications, e.g. citation network analysis~\cite{sun2009ranking}, recommendation systems~\cite{yu2014personalized} and drug-target interaction discovery~\cite{gonen2012predicting}. 
Fig.~\ref{fig1} provides a toy example of an HIN, which contains multiple types of nodes (e.g., Person and City) and edges (e.g., $BornIn$ and $LocatedIn$). The different types of nodes and edges entail different meanings and semantic relationships. To better describe the complex structure of HINs, we represent the two views of an HIN simultaneously as: 1) a \textit{schema graph}, which uses entity types and relations as nodes and edges to provide the meta-level description of the network, and 2) an \textit{instance graph}, which contains the instance-level observations of specific entities. 

Meta-paths are essential in analyzing HINs as they provide a higher-level description of the network structure that captures the relationships between different types of nodes and edges in the HINs.
Given two entities, a meta-path refers to a sequence of entity types and the relation that exists between them. For example, in Fig.~\ref{fig1}, the entity pair \textit{(Max Planck, Germany)} is connected via following meta-paths:

\noindent $\text{Person} \xrightarrow{BornIn} \text{City} \xrightarrow{LocatedIn} \text{Country}$ 

\noindent $\text{Person} \xrightarrow{GraduateFrom} \text{University} \xrightarrow{LocatedIn} \text{Country}$

\noindent $\text{Scientist} \xrightarrow{WorksAt} \text{University} \xrightarrow{LocatedIn} \text{Country}$

Collectively, these meta-paths indicate that \textit{Max Planck} is a citizen of \textit{Germany} and provide valuable insights 
% \sout{into the intricate relation between entities \textit{Max Planck} and \textit{Germany}}
into answering the query $(\textit{Max Planck}, \textit{isCitizenOf}, ? )$. 
% Such insights are valuable for understanding the underlying reasons behind the relationships between different entities. 
As meta-paths are effective in encapsulating interpretable semantics for relationships and transferring knowledge for reasoning about unseen entities~\cite{shi2016survey},
% \sout{Due to the strong power of meta paths in encapsulating explainable semantics for relations and transferring knowledge in reasoning unseen entities~\cite{shi2016survey},}
significant efforts have been made to leverage meta-paths for HIN link prediction tasks~\cite{sun2011co,yu2013recommendation}.

Notwithstanding the benefits of meta-paths, their applications in HINs are currently impeded by the challenges involved in their discovery. 
% \sout{Therefore, the complexity of meta-path discovery mainly arises from two factors: 1) the large meta-path search space and 2) the effective assessment of related meta-paths based on their path instance observations.} 
There are two main factors that pose challenges to the discovery of meta-paths in schema-complex HINs. The first entails the vast search space of potential meta-paths, and the second involves the burdensome task of accurately evaluating them through enumeration of path instances.
% the accurate evaluation of related meta-paths based on their path instance observations.
% 1) the large meta-path search space and 2) the lack of inductive meta-path learning methods, which can induce meta-paths for unqueried relations (i.e., relations whose meta-paths are not learned during training).
% \KC{unqueried relations can be vague. Explain it.}
% \KC{Our goal is to emphasize that we are the first to learn rules in an inductive setting.}
%3 why learning meta path is challenge (high complexity) 
% \KC{In this paragraph, we can summarize the existing works and point out it is challenging for them to handle complex HIN. To address the issue, we proposed an RL framework to handle it and briefly explain why RL is good at handling complex HIN.}

Firstly, the meta-path search space expands exponentially with the meta-path length. This expansion further occurs as the number of entity types and relations in the HIN increases. Therefore, 
% Firstly, the meta-path search space for an HIN with $|T|$ entity types and $|R|$ relations can potentially encompass $|T| \times (|R| \times |T|)^{l-1}$ candidate $l$-length meta-paths.
% Owing to this exponentially-growing search space, 
most existing meta-path-based approaches are only capable of handling schema-simple HINs. 
These are HINs with a limited number of entity types and relations (e.g., DBLP with 4 entity types~\cite{ley2002dblp}), where enumeration or manual design is feasible~\cite{gonen2012predicting,fu2017hin2vec,shi2018heterogeneous}.
% Given the large meta-path space, most existing meta-path-based methods could only deal with schema-simple HINs (e.g. DBLP~\cite{ley2002dblp} with 4 entity types), where enumeration or manual design is feasible~\cite{gonen2012predicting,fu2017hin2vec,shi2018heterogeneous}. 
Only a few works could reason schema-complex HINs such as knowledge bases (KBs) that cover a large number of entity types and relations (e.g., Dbpedia contains 174 entity types and 305 relations). 
% , to showcase the performance of their similarity functions
% For these work, some craft meta-paths with domain knowledge~\cite{gonen2012predicting} whilst others enumerate all possible meta-paths and select meta-paths in accordance with their needs~\cite{dong2017metapath2vec,fu2017hin2vec,shi2018heterogeneous}.
% As the latent meta-path space grows exponentially with the path length $l$ (theoretically $|T| \times (|T|\times|R|)^{l-1}$), high-quality meta-paths are computationally-prohibitive to obtain~\cite{lao2010pcrw,meng2015fspg}. 
% In contrast, the logical rule space is $|R|^{l-1}$, indicating the meta-path discovery problem is near $|T|^l$ times harder than the rule-learning problem in KBs.\KC{I would not recommend comparing meta-path learning and rule learning, as introducing too many concepts could potentially confuse the reader.} Even so, rule learning methods could hardly scale to KBs with hundreds of relations like Dbpedia. 
%where the number of entity types is relatively small (e.g., less than 10), and expert-crafted constructions are possible in these cases. 
%The number of all possible meta-paths grows exponentially with the path length~\cite{lao2010pcrw,meng2015fspg}, making it computationally-prohibitive to obtain high-quality meta-paths for various tasks.
%This might be the main reason why meta-paths are not as prevalent in knowledge bases as in schema-simple HINs. 
This intractable nature of meta-path discovery hampers the reasoning process, leading to their under-utilization in KB reasoning, especially when dealing with queries that involve a large number of relations.
% Most existing works for meta-path mining in schema-complex HINs are per-relation models, as they shrink the search space by only mining target relation-specific meta-paths during training. 
% \lsx{Most existing works separately learn independent models for each given query. For example, }
% The generated rules are only effective when reasoning the same target relation. 

% \KC{In this paragraph, we need to point out in additional to the difficulty in handling complex HIN. Existing methods cannot learn rule in inductive setting. Outline (1) what is rule learning in inductive setting. (2) Why it is challenging and why none of the existing methods can achieve it.}

Secondly, meta-paths are schema-level concepts, whose plausibility is evaluated based on the observation of paths at the instance level. 
% Since meta-path is a meta-level concept that cannot be directly observed in the instance graph, current methods heavily depend on path instances to discover meta-paths. 
For instance, as depicted in Fig. ~\ref{fig1}, to identify the meta-path $\text{Person} \xrightarrow{BornIn} \text{City} \xrightarrow{LocatedIn} \text{Country}$ in explaining \textit{isCitizenOf}, it is typically essential to refer to the evidence from corresponding path instances between \textit{isCitizenOf}-related entity pairs, such as $\text{Max Planck} \xrightarrow{BornIn} \text{Kiel} \xrightarrow{LocatedIn} \text{Germany}$.
% current methods could not effectively assess the quality of each meta-path. 
Sampling path instances of meta-paths could not guarantee effective evaluation whereas enumerating them is time-consuming.
% Existing works rely heavily on the observation of path instances as they adopt a bottom-top design that involves generating instance-level paths before summarization~\cite{meng2015fspg, wan2020mpdrl}.
Besides, as current methods heavily depend on path instances to discover meta-paths,
% as these methods rely heavily on the observation of path instances, 
they are incapable of handling inductive settings where the instance-level paths for a relation are unobserved. For example, existing methods cannot learn meta-paths to infer the query related to \textit{LivesIn} in Fig.~\ref{fig1} as there are no observed path instances for this relation. 
% \sout{These partial observations of path instances cannot guarantee the effective assessment of the meta-paths.} 
% \sout{Nevertheless, we claim that this assessment issue can be addressed using representation learning~\cite{cheng2022rlogic}.}

In this paper, we address aforementioned issues respectively by reinforcement learning (RL) and representation learning. The first issue of exploring the expansive meta-path space is tackled using RL, which offers an effective and flexible solution that can accommodate multiple training objectives for learning the meta-path discovery component.
To achieve this, we parameterize the discovery component as an RL-based path-finding agent provided after the evaluation of each meta-path. This agent learns to select plausible meta-paths by listening to rewards provided after meta-path evaluation.

For the meta-path evaluation issue, we directly represent meta-paths at the schema level by learning embeddings of entity types and relations~\cite{cheng2022rlogic}.
% Despite the challenges posed by inductive settings, we can overcome this issue through the use of representation learning.
% \sout{By representing meta-paths through representations of concepts and relations, we could learn meta-paths directly at the schema level, with scores calculated using the learned representations of concepts and relations.}
The plausibility of meta-paths can be subsequently assessed using the learned embeddings of entity types and relations.
% \KC{maintain consistency in using either "concepts" or "entity types" throughout the paper.}
% \sout{A small sample of meta-paths and their globally-evaluated scores would suffice for training the representations, eliminating the cumbersome necessity of finding all relevant path instances. Further, as some meta-paths could concurrently aid in explaining multiple relations (e.g., $\text{Person} \xrightarrow{BornIn} \text{City} \xrightarrow{LocatedIn} \text{Country}$ in explaining both the relations \textit{isCitizenOf} and \textit{LivesIn}), such dependency among their scores could be transferred via the schema-level representations.} 
Embeddings have a strong ability to capture the underlying similarities among entity types and relations for knowledge transfer, making them well-suited for addressing changes in inductive settings. For instance, even if we have not observed any path instances related to the relation \textit{LivesIn} in Fig.~\ref{fig1}, we can still infer the meta-path $\text{Person} \xrightarrow{isCitizenof} \text{Country} \xrightarrow{LocatedIn^{-1}} \text{City}$ for relation \textit{LivesIn}.
This inference is possible due to the similarity between the meta-paths explaining \textit{BornIn} (which has observed path instances) and \textit{LivesIn}, captured through the embeddings.
% due to the similarity between the meta-paths in explaining \textit{BornIn} and \textit{LivesIn}, which can be captured through embeddings. 
% In this sense, rather than solely relying on meta-paths instances for assessment, 
% Consequently, a meta-path can be evaluated without requiring the support from meta-path instances. 

We propose a novel method - \model, which can explore the large search space efficiently to perform a challenging task -  \textit{inductive meta-path learning} for the first time. 
In contrast to the instance-level inductive setting (a typical KB task) that aims to reason unseen entities, inductive meta-path learning requires learning meta-paths without finding the evidence of relation-specific path instances. 

The main contributions of this paper are as follows:
% \KC{it is better to emphasize our contribution in inductive rule learning }
\begin{enumerate}
    \item We make the first attempts to formalize an interesting problem -  inductive meta-path learning and proposed a novel framework to deal with complex HINs in the inductive setting.
    % \sout{discovery at schema level for schema-complex HINs, supporting our goal to learn meta-paths for multiple relations, especially in inductive meta-path learning scenarios} 
    % \KC{are we the first to frame the meta-path finding problem as a deterministic Markov Decision Process (MDP) on the schema graph? and why it is important to frame the meta-path finding problem directly on the schema graph. It may be more important to point out we are the first to learn rule in inductive setting.}
    
    \item We frame the meta-path learning problem as a Markov Decision Process (MDP) on the schema graph and design a novel RL-based agent, which jointly refines its policy and the schema-level representations to efficiently learn meta-paths of high coverage and confidence. During inference, the trained agent could promptly output/infer meta-paths for multiple relations with/without specific evidence of path instances. 

    \item We conduct extensive experiments, and demonstrate the superior ability of \model~in learning high-quality meta-paths for both KBs and general HINs. Impressively, \model~could effectively reason relations without finding specific evidence, with comparable performance to cases when these evidences for relations are provided.
\end{enumerate}

We discuss related work systematically in Section 2. 
%In Section 3, we first introduce definitions of meta-paths, coverage and confidence, and formulate the MDP of the meta-path finding problem. Afterwards, we describe \model~in detail, including the policy network architecture and the type node representations. 
In Section 3, after giving the necessary definitions and learning objectives, we formulate the MDP for \model, introduce the agent design, and outline the training pipeline. Section 4 presents the results in various settings, including multi-relation inductive setting, multi-relation transductive setting and per-relation transductive setting. 
% \sout{\model~is validated by query-answering and link prediction experiments and inductive studies, and compares \model~against state-of-the-art baseline methods on four real-world HINs.}
To validate \model, we conduct query-answering and link prediction experiments, as well as entity-level inductive studies. We also compare \model~against state-of-the-art baseline methods using four real-world HINs. We further analyze the convergence properties and the mined meta-paths in Section 5 and conclude the paper in Section 6. 

\section{Related Work}
\label{sec:related_work}
\noindent \textbf{Meta-path Reasoning.} After Sun et al. proposed meta-paths to carry semantic information and gauge entity relevance~\cite{sun2009ranking}, an increasing number of papers validated its applicability and performance in various tasks, including link prediction~\cite{sun2011co,  burke2014hybrid}. The most pivotal step for meta-path reasoning based approaches is to discover meta-paths.

For schema-simple HINs, the simplest way is through exhaustive enumeration up to a given length~\cite{wang2016relsim}, but is hideously computationally-expensive for schema-complex HINs. Some graph-traversal methods conduct breadth-first search~\cite{kong2012meta} or $A^{\ast}$ algorithm~\cite{zhu2018Aprime} on the network schema to generate meta-paths, however, the learning is hindered by the lacking of appropriate signal from the instance level. In particular, \textit{Autopath} 
combines deep content embedding and continuous RL to learn implicit meta-paths, and calculates the similarity scores as the empirical probabilities of reaching the target entities~\cite{yang2018autopath}.

For schema-complex HINs, the majority of current approaches involve a two-phase process that includes the generation and summarization of path instances. Lao and Cohen utilize random walks to generate meta-paths within fixed length $l$ and describe a learnable proximity measure with the discovered meta-paths, but $l$ is critical to the performance and varies greatly among datasets. \textit{FSPG} is a greedy approach that could consider user-input to derive the most relevant subset of meta-paths regarding the selected entity pairs~\cite{meng2015fspg}.
%, along with the \textit{GreedyTree} structure to facilitate the execution
%However, their model needs to pre-designate the source and target types (source type and target type are also generally identical) and could only deal with simple meta-paths. 
Wan et al. present \textit{MPDRL}, an RL method that considers type context to generate path instances, as an extension to \textit{MINERVA}~\cite{das2018minerva}.
%with lowest common ancestor principle. 
However, \textit{MPDRL} could be easily burdened by its path-finding component as its overall performance depends on partial observation (i.e., generated path instances).

\noindent \textbf{Multi-hop Reasoning.} Multi-hop reasoning methods leverage the information stored in the path instances connecting entity pairs~\cite{ji2021survey}. 
%To capture complex reasoning patterns, multi-hop reasoning methods, combinatory reason over the information stored in paths connecting entity pairs. 
These methods mainly disregard relation types and operate on inference paths obtained by random walks~\cite{lao2011random, guu2015traversing} or heuristics~\cite{toutanova2016compositional, gardner2014incorporating}. Inspired by the advancement of deep RL, Xiong et al. firstly formulate the path-finding process as sequential decision making and describe an RL framework for learning multi-hop relational paths~\cite{xiong2017deeppath}. Based on this, various papers explore different RL formulations and leverage different learning methods to boost multi-hop reasoning for prediction~\cite{das2018minerva,shen2018mwalk}.

\noindent \textbf{Embedding-based Reasoning.} Many KB completion tasks focus on learning vector representations for entities and relations with existing triples before using them for predictions. Most of these approaches ignore entity types and use different space for projection, e.g. \textit{TransE}~\cite{bordes2013transe}, \textit{KG2E}~\cite{he2015learning} and \textit{RotatE}~\cite{sun2019rotate}. However, these methods cannot capture compositional reasoning patterns and lack explainability ~\cite{guu2015traversing}. Recent works have also explored the embedding of HINs, premised on discovered meta-paths. Dong et al. propose \textit{Metapath2Vec}~\cite{dong2017metapath2vec} to embed HINs with crafted meta-paths, but its performance is restricted by limited existing meta-paths. In particular, \textit{HIN2Vec}~\cite{fu2017hin2vec} uses a multi-task learning approach to jointly embed the information of different relations and instance graph structure.

%Some recent works attempt to embed HINs with crafted meta-paths, e.g. \textit{Metapath2Vec}~\cite{dong2017metapath2vec} and \textit{HIN2Vec}~\cite{fu2017hin2vec}, yet their performance are restricted by limited number of existing meta-paths. Besides, all embedding based methods could bring blind-guess-level performance in inductive scenarios. 
\section{Methodology}
In this paper, we seek to learn meta-paths with high coverage and confidence. 
% We thus start this section by introducing the definitions of HIN, meta-paths, and the most commonly-used evaluation metrics for measuring their plausibility including coverage and confidence. 
We thus begin this section with an introduction to the definitions of HIN, meta-paths, and prevalent evaluation metrics, namely coverage and confidence, for assessing meta-path plausibility.
%Afterwards, we formulate the MDP for the meta-path finding problem. Finally, we present \model, including the policy network architecture and type node representations. 
To enable \model~to perform inductive meta-path learning, we place the path-finding agent on the schema graph with trainable schema-level representations. 
Therefore, following the definitions, we introduce \model~in detail by focusing on the MDP formulation for meta-path learning and the agent design. Finally, we present the training algorithm.

\subsection{Definitions and Objective}
\label{method:def}
In this subsection, we provide all necessary definitions before illustrating them with a guiding example. We then formulate the learning objective.
% \KC{the biggest problem of the section is that you define five concepts separately. But you didn't mention the relation among these five definitions. It confused the reader at the very beginning why we need these definitions and what is our goal? It is better to use one paragraph to define HIN (HIN can be further divided into schema graph $\mathcal{G}_S$ and instance graph $\mathcal{G}_I$ ) and meta path. Then mention our goal is to learn the meta path, where coverage and confidence is two widely used measure to evaluate meta path. after that, we use one example to illustrate what is HIN and how to compute coverage and confidence for a specific meta path.}

\textbf{Heterogeneous Information Network (HIN)}, denoted as $\mathcal{H} = (\mathcal{G}_I, \mathcal{G}_s)$ is a directed graph that encapsulates two perspectives: (1) a \textit{schema graph} for meta-level abstraction, denoted as $\mathcal{G}_s$, and (2) an \textit{instance graph} for instance-level instantiations, denoted as $\mathcal{G}_I$. In particular, $\mathcal{G}_I=(V, E)$ is formed with the set of entities denoted as $V$, and the set of links connecting entities denoted as $E\subseteq V\times V$. 
% \sout{with an entity type mapping $\tau$ and a relation mapping $\phi$, where $V$ denotes the entity set and $E\subseteq V\times V$ is the link set.} 
The type mapping $\tau:V\rightarrow2^T$ maps an entity to entity types, where $T$ denotes the entity type space; the relation mapping $\phi:E\rightarrow2^R$ labels a link to several relations, where $R$ is the relation set. The schema graph for HIN $\mathcal{H}$ is a directed graph $\mathcal{G}_S=(T,R)$ defined over entity types $T$, with links as sets of relations from $R$.
% \sout{In this paper, we refer to $\mathcal{G}_I$ as the \textit{instance graph} and name the network schema $\mathcal{G}_S$ as the \textit{schema graph}.}
% \sout{To provide a structured way to explore the complex relationships in the $\mathcal{G}_I$, the concept of meta-path is proposed~\cite{sun2011pathsim,meng2015fspg}, as follows.}
%\KC{the mapping function seems little wierd to me. We are supposed to map a set of entity to a set of entity type. But the mapping function map a set of entity to an integer}.
%\KC{We have several words to describe the same concepts: object types, entity type, type node....Please unify the terminology.}

A \textbf{meta-path} is a path defined on the schema graph of a HIN. It serves as a vital tool in HIN analysis, capturing the semantic relationships between different types of entities in the network.
A meta-path $M$ of length $l$ is defined as $M=t_1\xrightarrow{r_1}t_2\xrightarrow{r_2}\cdots\xrightarrow{r_{l-1}}t_l $, where $t_i\in T$ denotes an entity type and $r_i\in R$ represents a relation. An instance-level path $P=v_1\xrightarrow{r_1}v_2\xrightarrow{r_2}\cdots\xrightarrow{r_{l-1}}v_l$ satisfies $M$ if $\forall i\in \{1,\cdots,l\}, t_i\in \tau(v_i)$ and $\forall i\in \{1,\cdots,l-1\}, r_i\in \phi(v_i, v_{i+1})$. In this case, we also say $P$ is a path instance specific for $M$.
% , and entity pair $(v_1, v_l)$ is connected via $M$. 
% \sout{A meta-path is valid if there exists any meta-path instance for it, in $\mathcal{G}_I$.} 
%A HIN $\mathcal{H}$ is a directed graph $G=(V, E)$ with a type mapping $\tau$ and a relation mapping $\phi$. $V$ denotes the set of entities on the graph and $E\subseteq V\times V$ is the set of edges connecting entities in $V$. The type mapping is given as $\tau:V\rightarrow2^T$, where $T$ is the set of node types; the relation mapping is $\phi:E\rightarrow2^R$, where $R$ is the set of relations. 
% \noindent\textbf{Definition 2 (Network Schema)} 
%\KC{Every HIN has an schema graph $T_G=(T,R)$ defined over entity types $T$ and relations $R$.} 
%\KC{when we define Heterogeneous Information Network, please also define instance graph and schema graph.}\FC{Agree with Vivian}
%There are also other meta-paths summarizing the paths connecting {Max Planck} and {Germany}. 
%\KC{meta-paths is a schema level concept. We can summarize the instance level paths in order to discover meta-paths. But meta-paths won't summarize the instance level observations (i.e, paths connecting {Max Planck} and {Germany})  }
% \sout{Meta-path assessment plays a crucial role in meta-path learning, allowing us to evaluate the quality of the learned meta-paths.}
The usage of meta-path instances is widespread for assessing the plausibility of meta-paths. In the field of association rule mining, coverage and confidence are two of the most crucial metrics for rule evaluation, each offering a distinct perspective~\cite{agrawal1993mining}.
Coverage denotes the rule's applicability, while confidence serves as a gauge for its reliability. Here, we delve deeper these concepts and establish the rationale for combining these two metrics.
% Together, they provide a comprehensive measure of the usefulness of a meta-path.
% \sout{There are two widely-used metrics for this purpose: coverage and confidence.} 
% To evaluate the quality of meta-paths, coverage and confidence are two widely-used metrics.
% , based on (observed) path instances.

% \noindent\textbf{Coverage. } 
The \textbf{coverage} of a meta-path $M$ for a specific relation $r_q$ quantifies the frequency at which the meta-path occurs in $r_q$-related entity pairs. 
It is calculated by taking the ratio of entity pairs connected by both $r_q$ and $M$, to the total number of $r_q$-connected entity pairs,

\begin{equation}
% \footnotesize
\resizebox{\hsize}{!}{$Cvrg_{M\Rightarrow r_q}^\mathcal{H}=\frac{|\{(v_i,v_j)|Con_M(v_i,v_j),r_q\in \phi(v_i, v_j), v_i,v_j\in \mathcal{H}\}|}{|\{(v_i,v_j)|r_q\in \phi(v_i, v_j), v_i,v_j\in \mathcal{H}\}|}$}
\label{equ_cover}
\end{equation}
where $Con_M(v_i, v_j)$ indicates a meta-path instance that connects the entity pair $(v_i, v_j)$.

The \textbf{confidence} of a meta-path $M$ in accurately identifying $r_q$ evaluates the trustworthiness of the associations observed along the meta-path to provide an explanation for $r_q$. 
This metric is computed as the conditional probability of observing the consequent entities in $\mathcal{G}_I$ given the presence of the antecedent entities along the meta-path $M$, as given below. The denominator signifies the total number of entity pairs connected by the meta-path $M$:

\begin{equation}
% \footnotesize
\resizebox{\hsize}{!}{$Conf^\mathcal{H}_{M\Rightarrow r_q}=\frac{|\{(v_i,v_j)|Con_M(v_i,v_j),r_q\in \phi(v_i, v_j), v_i,v_j\in \mathcal{H}\}|}{|\{(v_i,v_j)|Con_M(v_i,v_j), v_i,v_j\in \mathcal{H}\}|}$}
\label{equ_conf}
\end{equation}

% Rules can exhibit high accuracy but limited applicability (high confidence, low coverage), or broad applicability but unreliable consistency (high coverage, low confidence). Integrating these metrics identifies rules exhibiting both applicability and accuracy. 

Given the entity pairs connected by a relation, coverage identifies the frequency of a meta-path being satisfied by the paths connecting these entity pairs whilst confidence measures the correctness of a meta-path's representation of the relation. 
Meta-paths could display high accuracy but limited applicability (high confidence, low coverage), or broad applicability but low accuracy (high coverage, low confidence). By integrating these two metrics, we could identify rules that are of greater relevance and utility. 
% identifies rules exhibiting both applicability and accuracy. 
% A meta-path exhibiting high levels of both coverage and confidence is likely to be of greater relevance and utility.
The complexity in computing these metrics for a meta-path involves finding all its path instances (also known as the evidence for it).

\noindent \textbf{Example. }Fig. \ref{fig1} shows a toy example of HIN, with $T=\{Person, Scientist, University, City, Country\}$ and $R=\{isCitizenof, GraduatedFrom, WorksAt, LocatedIn\\BornIn, LivesIn\}$. The relations between \textit{Max Planck} and \textit{Germany} is $\phi(Max Planck, Germany)=\{isCitizen\}$ and {Max Planck} is mapped into types: $\tau(Max Planck)=\{Person, Scientist\}$. To reason the relation \textit{isCitizenOf}, we could refer to following meta-path:

\noindent $Person\xrightarrow{GraduatedFrom}University\xrightarrow{LocatedIn}Country$

which is satisfied by the instance path $Max Planck\\ \xrightarrow{GraduatedFrom}Univ. of Munich\xrightarrow{LocatedIn}Germany$. The entity pair $(Max Planck, Germany)$ is thereby connected via such meta-path. If the HIN indicates following information: 250 people graduated from German universities, among them 150 are German citizens and 100 are international students, and there are 200 German citizens in total.
In this sense, the coverage of the above meta-path with respect to relation \textit{isCitizenOf} is $\frac{150}{200}$, and the confidence is $\frac{150}{200+100}$. 
% \KC{it is better to use the example in the figure even though the number is small (e.g., 1/2,0) }
% If we further add information on citizens of other countries along with their alma mater to the HIN, we need to examine the meta-path connectivity and relation existence for each entity pair when calculating \emph{coverage} and \emph{confidence}. In practice, this is done by calculating the product of adjacency matrices, which indicates the connectivity of entity pairs, for relations on the meta-path or query relation, and is accelerated by harnessing the sparse matrix data structure.

\noindent \textbf{Learning Objective.} In this paper, we aim to learn meta-paths with high coverage and confidence for relations directly at the schema level.
% \sout{, without requiring evidence specific for these relations.} 
% \sout{To achieve this, we generate meta-path samples for some training relations, evaluate their scores based on specific evidence, and refine the generation policy and schema-level representations. }
To achieve this goal, we train a path-finding agent on the schema graph to effectively navigate and explore the most promising meta-paths that lead to the relations. The evaluation of these meta-paths is performed using meta-path instances found in the instance graph.
% \sout{During test, the model infers meta-paths for some untrained test relations. We refer to this as the \textit{multi-relation inductive} setting (Fig. \ref{fig_setting} (A)). Since the trained and tested relations are the same in typical KB reasoning, we add it as the \textit{multi-relation transductive} setting (Fig. \ref{fig_setting} (B)). Additionally, we include the traditional HIN reasoning task, which trains and tests over a single relation as the \textit{per-relation transductive} setting (Fig. \ref{fig_setting} (C)). }
To demonstrate the effectiveness of our proposed methods across various scenarios, we assess the performance of our learned model using three distinct settings, as illustrated in Fig. \ref{fig_setting}. In the \textit{multi-relation inductive setting} (Fig. \ref{fig_setting} (A)), a single model is trained to discover meta-paths for multiple relations. The relations used in training are different from the test dataset, ensuring a disjoint relationship between them. Consequently, the meta-path instances for the relations in the test dataset are unseen during the training phase. Due to the inherent challenges in the multi-relation inductive setting, many existing methods struggle to effectively address it. To comprehensively evaluate our proposed methods, we include two additional settings that are comparatively less challenging. In the \textit{multi-relation transductive setting} (Fig. \ref{fig_setting} (B)), a single model is trained to discover meta-paths for multiple relations. The target relations used for training the model encompass the relations present in the test dataset. In the \textit{per-relation transductive setting} (Fig. \ref{fig_setting} (C)), a separate model is trained for each single relation to discover specific meta-paths exclusively tailored for that particular relation.

% \sout{In fact, a given relation $r_q$ could connect multiple type pairs on the $\mathcal{G}_S$. For example, the relation $isCitizenof$ connects both $(Person, Country)$ and $(Scientist, Country)$. Each $r_q$-related type pair $(t_{\rm src}, t_{\rm tgt})$ can be combined with $r_q$ to form a query $r_q(t_{\rm src}, t_{\rm tgt})=?$. To best explain $r_q$, we wish to find rewarding meta-paths for all such $r_q$-related queries. In multi-relation settings, we would adopt all queries for all the target relations.}\KC{this part is unclear to me. didn't get the reason why we want to mention it.}

% During training, we generate meta-paths for all relevant queries and assess them to update the meta-path generator, and we infer meta-paths with the trained generator during test. Once a query is used for training, it becomes an observed query.
% In the inductive meta-path learning setting, we infer meta-paths for unobserved queries that corresponds to unqueried relations, during test.
% To reason a certain relation $r_q$, we aim to find a set of meta-paths with high coverage and confidence to explain it. 

\begin{figure*}[htb]
\centering
\includegraphics[width=\linewidth]{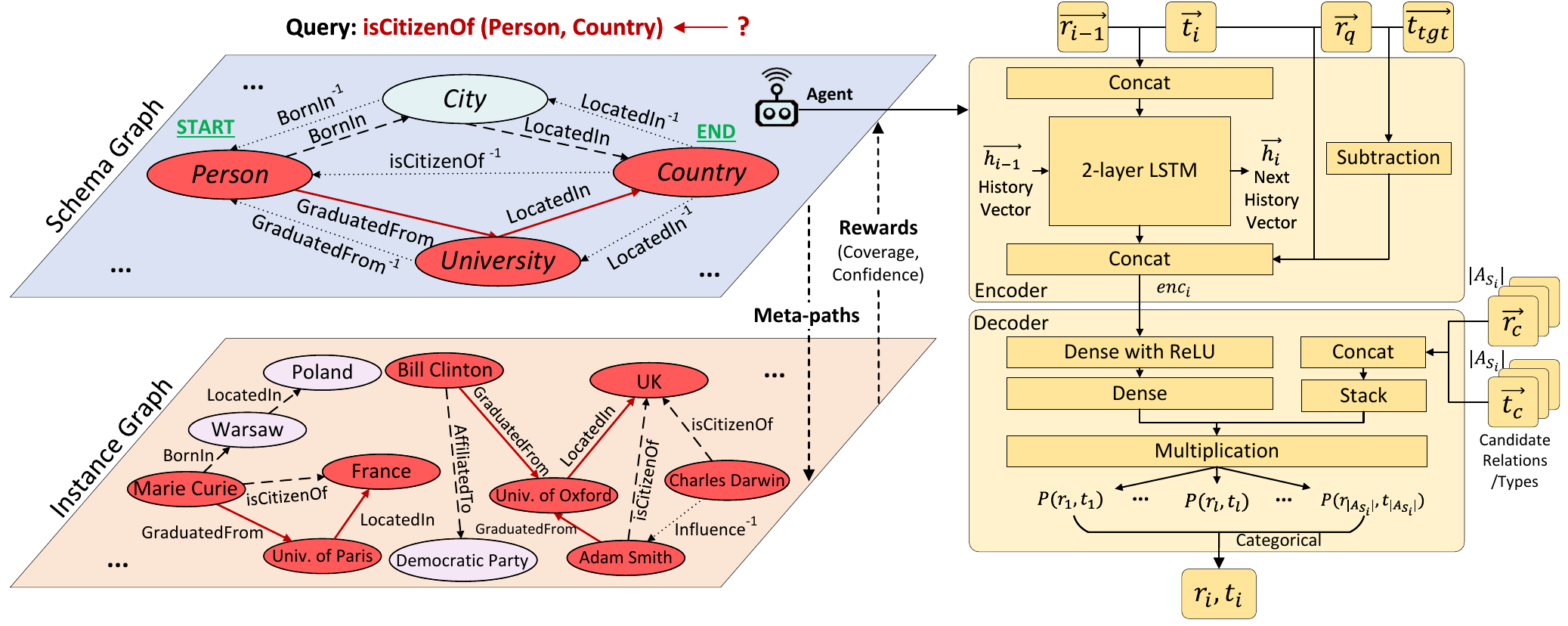}
\caption{Overview of \model. The \textbf{Left} part describes the interactions between the two views of HIN. The upper schema graph provides the learning environment where the agent is trained to navigate to the target type node based on a query, e.g. \textit{isCitizenOf(Person, Country)=?}, and establishes meta-paths, e.g. $Person \xrightarrow{GraduatedFrom} University \xrightarrow{LocatedIn} Country$. The lower instance graph provides the rewards: the coverage and confidence of the discovered meta-path are calculated based on the instance paths satisfying the meta-path, e.g. $Marie Curie \xrightarrow{GraduatedFrom} Univ. of Paris \xrightarrow{LocatedIn} France$. Subsequently, the rewards would be utilized to train the agent. Note that the red arrows denote the discovered meta-path and related instance-paths, dashed arrows denote existing relations, and dotted arrows indicate the inverse ones. 
% In inductive meta-path learning settings, the agent needs to infer meta-paths for unobserved queries (e.g., \textit{BornIn(Person, City)=?}) during test.
The \textbf{Right} part shows the detailed encoder-decoder based policy network architecture.}\label{fig2}
\end{figure*}

\subsection{Reinforcement Learning Formulation}\label{method:mdp}
As discussed in Section~\ref{sec:related_work}, current meta-path discovery methods adopt a bottom-top design. For example, MPDRL employs RL to navigate the instance graph, creating path instances prior to summarizing them as meta-paths. 
% In addition to facing issues with partial observation, this design also prohibits inductive meta-path learning due to the absence of schema-level representations.
This approach not only faces challenges associated with the introduction of bias through path instances sampling but also prevents the direct learning of rules at the schema level.
To fully unleash the expressive power of schema-level representations, we formulate the meta-path generator as a path-finding agent on the schema graph.
% \KC{Although we can define RL framework clearly, we didn't give insight into why this framework is the best choice to model meta path learning problems. emphasize we aim to handle the complex HIN and we need efficient exploration in searching space.}
% For meta-path learning, since 1) the meta-path space is gigantic and 2) the time complexity mainly originates from meta-path assessment, we adopt RL-based agent for efficient exploration.
Given the extensive meta-path space, we adopt an RL-based agent for efficient exploration.
% The meta-path learning problem, as a Markov Decision Process (MDP) on the schema graph, is thus given as below. 
As depicted in Fig. \ref{fig2} (Left), each time for the relation $r_q$, the agent randomly starts at a source type $t_{\rm src}$ in the entity type pairs $(t_{\rm src}, t_{\rm tgt})$ connected by $r_q$ on the $\mathcal{G}_S$.
This forms the query $r_q(t_{\rm src}, t_{\rm tgt})=?$ for the agent who dynamically adjusts its policy to arrive at $t_{\rm tgt}$ to form meta-paths and simultaneously maximizing the rewards for the discovered meta-paths. We thus describe the meta-path learning problem, as a Markov Decision Process (MDP) on the schema graph by the tuple $(\mathcal{S}, \mathcal{A}, \mathcal{P}, \mathcal{R})$.

% \KC{The motivation of defining state, actions, reward in this way. For example, why do we use confidence and coverage for reward? How does the state cover all the relevant information we need? Why this information is important in this scenario. }

\noindent\textbf{States}. Intuitively, we wish the agent to have sufficient knowledge of its current position and all information about the query ($r_q(t_{\rm src}, t_{\rm tgt})=?$) during its navigation. Therefore, at step $i$, the state $S_i$ is represented by $(t_i, t_{\rm src}, r_q, t_{\rm tgt})$, where $t_i$ stands for the current entity type node. The state space $\mathcal{S}$ contains all valid combinations in $T\times T\times R\times T$.

\noindent\textbf{Actions}. Given a state $S_i=(t_i, t_{\rm src}, r_q, t_{\rm tgt})$, the action space $A_{S_i}$ contains all outgoing edges of the type node $t_i$ in schema graph $T_G$ along with an operation to stay unmoved, i.e. $A_{S_i}=\{(r, t)|(t_i,r,t)\in T_G\}\cup\{(\emptyset, t_i)\}$. Starting at node $t_{\rm src}$, the agent selects the most favoured action for $l-1$ times based on the representations of edge $r$ and outgoing node $t$, forming a $l$-length meta-path or halting at $t_{\rm tgt}$ if it reaches the target node $t_{\rm tgt}$ before the maximum hop.

\noindent\textbf{Transition.} The environment dynamics is characterized by the state transition probability $\mathcal{P}:\mathcal{S}\times \mathcal{A}\rightarrow \mathcal{S}$. Given $S=(t_i,t_{\rm src},r_q,t_{\rm tgt})$ and $A=(t_i,r,t)$, then $\mathcal{P}(S,A)=(t,t_{\rm src},r_q,t_{\rm tgt})$, implying that the state is updated to the new type node $t$ via the chosen $r$.

\noindent\textbf{Rewards.}
As meta-paths are a concept at the schema level, evaluating their performance necessitates incorporating evidence from instance graphs. To achieve this, we integrate the commonly used metrics of coverage and confidence, based on meta-path instances, into the reward function. Moreover, we introduce an arrival indicator $\mathbb I_{\rm arr}(M) \in \{0,1\}$ to guide the navigation process, as outlined below:
% \sout{Since we wish to learn meta-paths of high coverage and confidence, we incorporate them into the reward function. In addition, we add an arrival indicator $\mathbb I_{\rm arr}(M) \in \{0,1\}$ to guide the navigation, as follows,}
%It is comprised of two parts: first, coverage in the instance graph (Eq.~(\ref{equ_cover})) and second, the arrival indicator  $\mathbb I_{\rm arr}(M) \in \{0,1\}$ defined as follows:
% comprised of two parts. First is the coverage, as defined in Eq.~(\ref{equ_cover}), second is the arrival indicator as follows,
\begin{equation}
% \footnotesize
 \resizebox{\hsize}{!}{$\mathbb I_{\rm arr}(M)=\mathbb I \{t_l=t_{\rm tgt}\} \; \& \;  \mathbb I \{\exists\  i\in \{1,\cdots,l-1\},  A_i\neq (\emptyset,t_i)\}$}
%\mathbb I_{\rm arrival}(M)=\mathbb I \{t_l=t_{\rm tgt}\} \; \& \;  \mathbb I \{t_i\neq (t_{\rm src}), \exists i\in \{1,\cdots,l-1\}\}
\label{equ_arrival}
\end{equation}
% \begin{equation}
% \mathbb I_{\rm arrival}(M)=\mathbb I \{t_l=t_{\rm tgt}\} \; \& \;  \mathbb I \{ t_i\neq t_{\rm src}, \exists i\in \{1,\cdots,l\}\}
% \label{equ_arrival}
% \end{equation}

\noindent where $\&$ denotes the logical-and operation. The arrival indicator ensures the agent reaches the target type node $t_{\rm tgt}$ (first term) and prevents the agent from sticking at the starting node $t_{\rm src}$
%or encourage the agent samples longer paths rather not just one hop from source to target
(second term). 
% The second term is useful when we reason relations that could connect the same type node, e.g., $Team\xrightarrow{PlaysAganist}Team$. 
The second term is useful when the agent observes queries with the same source and target type node, such as $PlaysAganist(Team, Team)=?$. 
Otherwise, the agent may mistakenly believe that it has reached the target type ($Team$) as it starts and therefore never move, rendering no meta-paths.
The three parts are combined with different weights, resulting in normalized rewards in the range $[0,1]$ as follows:
\begin{equation}
% \footnotesize
\resizebox{\hsize}{!}{$R(M|r_q,\mathcal{H})=\frac{\lambda_1* Cvrg^\mathcal{H}_{M\Rightarrow r_q}+\lambda_2*Conf^\mathcal{H}_{M\Rightarrow r_q}+\mathbb I_{\rm arr}(M)}{\lambda_1+\lambda_2+1}$}
\label{equ_reward}
\end{equation}
%\lsx{Move to Appendix: We set $\lambda=2$ due to empirical performances.}
% \vspace{-2pt}
\subsection{Agent Design}
\vspace{-2pt}
\label{method:agent}
% \KC{emphasize why we can handle inductive setting. From my understanding, because we learn the representation of relations and we have a function to model the decision-making process. Thus, we can handle inductive rule learning.}
With the above formulation, the agent learns to navigate the schema graph by listening to the reward signal. However, such formulation alone could not support meta-path learning in the absence of specific evidence. Consequently, we incorporate representation learning to further empower the search ability and, more importantly, encapsulate the underlying similarities among entity types and relations.
The latter is particularly useful when the agent learns to navigate based on an unobserved query, as in inductive meta-path learning settings.
Specifically, we design the policy network with an encoder-decoder architecture that incorporates schema-level representations. In this section, we denote vector representations with bold fonts.
%\lsx{Delete: We now introduce the policy network of \model~in detail, which is designed following an encoder-decoder architecture.}
\begin{figure*}[h!]
\centering
\includegraphics[width=\linewidth]{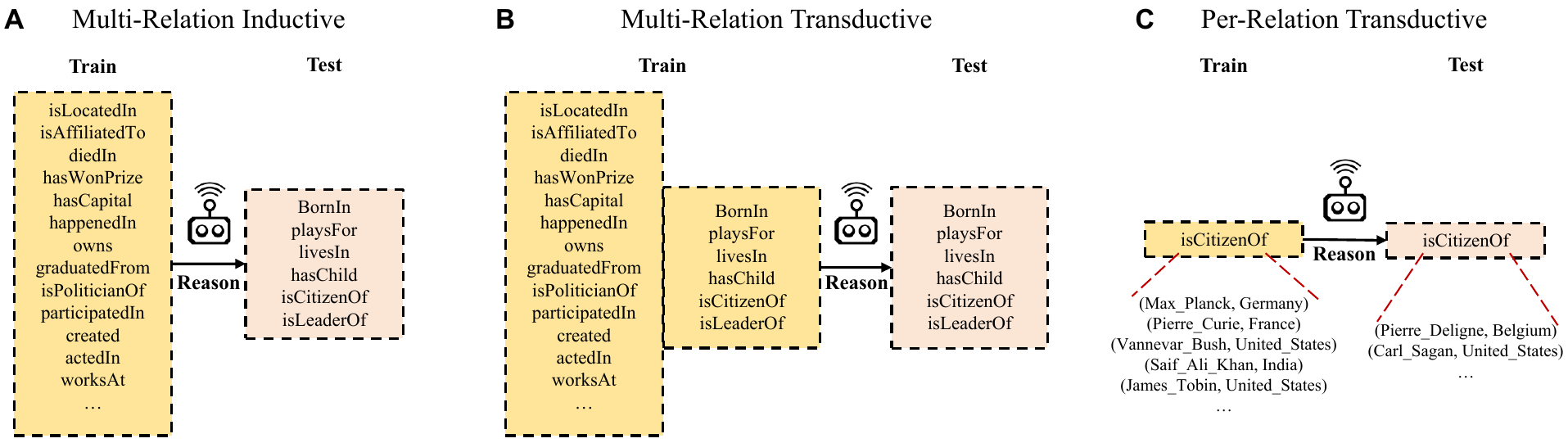}
\caption{Three experimental settings adopted in this paper (exemplified using the relations and entity pairs in YAGO26K-906). 
% \KC{good example. it is better to show the inductive setting in Fig 2. For example change the query in fig 2 to an unobserved query.}
}\label{fig_setting}
\end{figure*}

\noindent\textbf{Policy Network. } 
%\lsx{Deleted: To search meta-paths in schema-complex HINs efficiently, we craft a policy network with an encoder-decoder design.} 
As illustrated in Fig. \ref{fig2} (Right), the policy network adopts an encoder-decoder architecture. The encoder is parameterized by a two-layer long short term memory network (LSTM), which encodes the state $S_i$ at each time step $i$ into a vector representation $\mathbf{S}_i$. 
% Although alternative models, such as Transformers, could potentially be employed, we opted for LSTM due to its efficient processing of short sequences and because other structures did not yield a significant performance enhancement.
While other models like Transformers could have been considered, we opted for LSTM due to its fewer parameters, enabling rapid convergence while yielding satisfactory performance, even with limited sequence data.
% While other models like Transformers could have been considered, we opted for LSTM due to its efficient processing of short sequences.
Specifically, the LSTM takes the concatenation of the vector representations of relation $\mathbf{r}_{i-1} \in \mathbb{R}^{d_e}$ and type node $\mathbf{t}_i\in \mathbb{R}^{d_e}$ as inputs and outputs $\mathbf{S}_i$ and the updated history vector $\mathbf{h}_i$.
To form the encoding $\mathbf{enc}_i$, $\mathbf{S}_i$ is concatenated with $\mathbf{t}_i$, $\mathbf{r}_q$, and the difference between $\mathbf{t}_{\rm tgt}$ and $\mathbf{r}_q$. 
% $\mathbf{S}_i$ is further concatenated with $\mathbf{t}_i$, $\mathbf{r}_q$ and the difference between $\mathbf{t}_{\rm tgt}$ and $\mathbf{r}_q$, to form the encoding ${\bf enc}_i$.
The design of the difference item helps the agent find the adjacent head types for the query's target type. We use $[\cdot \mathbin\Vert \cdot]$ to represent the concatenation operator.

\begin{equation}
% \footnotesize
\begin{split}
&{\bf h}_i, {\bf S}_i={\rm LSTM}({\bf h}_{i-1}, [{\bf r}_{i-1}\mathbin\Vert {\bf t}_i])
\\
&{\bf enc}_i=[{\bf S}_i\mathbin\Vert {\bf t}_i \mathbin\Vert {\bf r}_q \mathbin\Vert ({\bf t}_{\rm tgt}-{\bf r}_q)]
\label{equ_enc}
\end{split}
\end{equation}

\noindent The decoder is implemented by a two-layer multi-layer perception (MLP) network (whose hidden size is denoted as $d_h$) with rectified linear unit. All the candidate relations $r_c$ and type nodes $t_c$ are determined by the outgoing edges of the current type node. 
Each edge is represented by the concatenation of $\mathbf{r}_c$ and $\mathbf{t}_c$. These representations are stacked to create a feature matrix for decision $D_i$.
% We represent an edge by the concatenation of $\mathbf{r}_c$ and $\mathbf{t}_c$ and stack all the representations to obtain the feature matrix for decision $D_i$. 
After passing the encoding ${\bf enc}_i$ to the MLP, the output is multiplied by $D_i$ and then it goes through a softmax layer to calculate the action probability ${\bf P}\in \mathbb{R}^{|A_{S_i}|}$,

\begin{equation}
% \footnotesize
\begin{split}
&D_i=\circ([{\bf r_c} \mathbin\Vert {\bf t_c}], c=\{1,...,|A_{S_i}|\})
\\
&{\bf P}={\rm softmax}(D_i(W_2({\rm ReLU}(W_1{\rm enc_i}+b_1))+b_2)
\label{equ_dec}
\end{split}
\end{equation}
\noindent where, $\circ(\cdot)$ is the stacking operator. The overall policy network is represented by $\pi_\theta$, where $\theta$ corresponds to the parameters in the LSTM and MLP. Based on the action probability distribution $\mathbf{P}$, the agent selects an action and subsequently moves to the next type node with regard to the transition dynamics.

\subsection{Training Pipeline}
\label{method:train}
Due to the complex nature of meta-path finding, we leverage RL algorithm to train the policy network $\pi_\theta$ of the agent to maximize the rewards described in Eq.~(\ref{equ_reward}).
To support multi-relation reasoning, we employ a periodic design that involves switching the relation for training every $I_{base}$ iterations, enabling each relation to undergo $I_r$ rounds of training.
The training objective is,  
\begin{equation}
J(\theta|r_q, \mathcal{H})=\mathbb{E}_{(t_{\rm src}, r_q, t_{\rm tgt})\sim \mathcal{H}}\mathbb{E}_{M\sim \pi_\theta(M)}[R(M|r_q,\mathcal{H})]
\end{equation}
where, $\pi_\theta(M)$ is the distribution for the generated $l$-length meta-path following policy $\pi_\theta$. Following the REINFORCE algorithm~\cite{williams1992simple,sutton2000policy}, the objective could be optimized in the direction of,

\begin{equation}
\footnotesize
\begin{split}
\nabla_\theta &J(\theta|r_q, \mathcal{H})=\mathbb{E}_{(t_{\rm src}, r_q, t_{\rm tgt})\sim \mathcal{H}}\mathbb{E}_{M\sim \pi_\theta(M)}[R(M|r_q,\mathcal{H})\nabla_\theta \ln \pi_\theta(M)]
\\
&=\mathbb{E}_{(t_{\rm src}, r_q, t_{\rm tgt})\sim \mathcal{H}}\mathbb{E}_{M\sim \pi_\theta(M)}[R(M|r_q,\mathcal{H}) \sum_{i=1}^l\nabla_\theta \ln \pi_\theta(A_i|S_i)]
\label{equ_gradient}
\end{split}
\end{equation}

% \begin{equation}
% \begin{split}
% \nabla_\theta J(\theta|r_q, \mathcal{H})=\mathbb{E}_{(t_{\rm src}, r_q, t_{\rm tgt})\sim \mathcal{H}}\mathbb{E}_{M\sim \pi_\theta(M)}[R(M|r_q,\mathcal{H})\nabla_\theta \ln \pi_\theta(M)]
% \\
% =\mathbb{E}_{(t_{\rm src}, r_q, t_{\rm tgt})\sim \mathcal{H}}\mathbb{E}_{M\sim \pi_\theta(M)}[R(M|r_q,\mathcal{H}) \sum_{i=1}^l\nabla_\theta \ln \pi_\theta(A_i|S_i)]
% \label{equ_gradient}
% \end{split}
% \end{equation}

\noindent To estimate the gradient in Eq.~(\ref{equ_gradient}), we randomly draw $K$ entity type pair $(t_{\rm src}, t_{\rm tgt})$ connected by $r_q$ from $\mathcal{H}$ and run $N$ roll-outs for each sample. The gradient is approximated by the sampled trajectories as follows,

\begin{equation}
\footnotesize
\nabla_\theta J(\theta|r_q, \mathcal{H})\approx \frac{1}{N\cdot K}\sum_{j=1}^{N\cdot K}[R(M|r_q,\mathcal{H}) \sum_{i=1}^l\nabla_\theta \ln \pi_\theta(A_i^j|S_i^j)]
\label{equ_approx}
\end{equation}

\noindent We append a moving average baseline to stabilize the training process by reducing the variance. The baseline is calculated by averaging the accumulative discounted rewards. 
%We abandon the Actor-Critic pipeline with a parametric baseline~\cite{konda2000actor} due to the poor practical performances.
Although the Actor-Critic pipeline with a parametric baseline prevails in the field of RL~\cite{konda2000actor}, we do not observe statistical significance by adopting it. Besides, to spur the exploratory behavior of \model~in discovering diverse meta-paths, 
%To encourage \model~explore more diverse meta-paths, 
we include an entropy regularization term with weight $\beta$ (with a decay rate) in the loss function.
%~\cite{mnih2016asynchronous}. %Adam gradient descent updates on mini-batch samples are performed to minimize
Ultimately, the ADAM optimizer is used to minimize the loss by rate $\alpha$. With above training process, the policy network $\pi_\theta$ guides the learning of trainable representations of entity types and relations with the goal of maximizing the reward.
% With above training process, the policy network $\pi_\theta$ undergoes appropriate adjustments, leading to updates in the encoding as well as the trainable representations of entity types and relations that contribute to the formation of encoding.
\section{Experiments} 
\begin{table}[htb]
\centering
\caption{Categories of baseline methods}
\resizebox{\linewidth}{!}{
\begin{tabular}{c|c|c|c}
\toprule[1.5pt]
Names & Meta-path-based   & Embedding-based  & Rule-learning                               \\ \hline
\textbf{RotatE}                 &         & \checkmark       &                  \\ 
\textbf{TransE}                 &         &     \checkmark   &          \\ 
\textbf{DistMult}                 &         &     \checkmark   &          \\ 
\textbf{ComplEx}                 &         &     \checkmark   &          \\ 
\textbf{Random Walk}                  &     \checkmark    &        &                              \\ 
\textbf{MPDRL}                  &     \checkmark    &        &                              \\ 
\textbf{PCRW}                   &     \checkmark    &        &                           \\ 
\textbf{Autopath}               &        \checkmark &        &  \checkmark            \\ 
\textbf{Metapath2Vec}           &    \checkmark     &  \checkmark      &                    \\ 
\textbf{HIN2Vec}                &   \checkmark      &   \checkmark     &           \\ 
\textbf{NHSE}                &         &   \checkmark     &                 \\
\textbf{RNNLogic}                 &         &        &   \checkmark       \\ 
\textbf{MINERVA}                &         &        &  \checkmark               \\
\textbf{MLN4KB}                &         &        &  \checkmark               \\
\bottomrule[1.5pt]

\end{tabular}
}
\label{tab:baseline}
\end{table}
In this section, we present the empirical experimental results to establish following points: (i) \model~can effectively and efficiently generate meta-paths without relation-specific evidence (multi-relation inductive setting, as in Sec. \ref{sec:experiment:multi-relation_inductive}), (ii) \model~is competitive against state-of-the-art KB reasoning baselines in the query answering task (multi-relation transductive setting, as in Sec. \ref{sec:experiment:multi-relation_transductive}), (iii) \model~is effective in reasoning general HINs (per-relation transductive setting, as in Sec. \ref{sec:experiment:per-relation_transductive}). The distinctions between the three settings are illustrated in Fig. \ref{fig_setting}. In the per-relation transductive setting (which is prevalent in the current HIN reasoning experiments), the agent is trained and tested over entity pairs connected by a specific relation in each run. For the two multi-relation settings, we train and test our model on entity pairs associated with multiple relations. In the multi-relation inductive setting, the trained and test relations do not overlap. Overall, we compare against the baselines shown in Table \ref{tab:baseline}.

% \begin{table*}[htb]
%     \centering
%     \renewcommand\arraystretch{1.13}
%        \caption{Inductive KG Query-answering results on YAGO26K-906 and DB111K-174, averaged over 5 runs. Enumeration cannot scale to DB111K-174 and hence the results are not reported.}
% \resizebox{0.95\linewidth}{!}{%
%  \begin{tabular}{cccccccc}
%     \toprule[1.5pt]
%    & Hits@1 & Hits@3 & Hits@10
%    & MRR & \#Output Metapaths & \#Valid Metapaths & Valid Rate 
% \\
%      \hline
%      \multicolumn{8}{c}{YAGO26K-906}
% \\
% \model&$0.146$&$0.239$&$0.376$&$0.218$&$20,326$&$2,687$&\pmb{$13.22\%$} 
% \\
% Random Walk&$0.066$&$0.097$&$0.139$&$0.089$&$19,037$&$387$&$2.03\%$ 
% \\
% Random Walk (*5)&$0.114$&$0.169$&$0.220$&$0.151$&$92,003$&$1,561$&$1.70\%$ 
% \\
% Random Walk (*10)&$0.121$&$0.191$&$0.288$&$0.174$&$187,400$&$2,940$&$1.57\%$ 
% \\
% Enumeration&$0.312$&$0.452$&$0.564$&$0.400$&$14,894,588$&$152,502$&$1.02\%$ 
% \\
% \hline
%      \multicolumn{8}{c}{DB111K-174}
% \\
% \model&$0.638$&$0.825$&$0.851$&$0.731$&$30,847$&$979$&\pmb{$3.17\%$}
% \\
% Random Walk&$0.010$&$0.014$&$0.017$&$0.012$&$22,506$&$111$&$0.49\%$ 
% \\
% Random Walk (*5)&$0.032$&$0.048$&$0.062$&$0.043$&$111,892$&$448$&$0.40\%$ 
% \\
% Random Walk (*10)&$0.070$&$0.105$&$0.137$&$0.093$&$222,143$&$788$&$0.35\%$ 
% \\
% Enumeration&-&-&-&-&-&-&-
% \\
% \bottomrule[1.5pt]
% \end{tabular}
% }
% \label{table_multi_inductive}
% \end{table*}

\begin{table*}[htb]
    \caption{Inductive KG reasoning results on YAGO26K-906 and DB111K-174, averaged over 5 runs. Enumeration cannot scale to DB111K-174 and hence the results are not reported. The best/second best metrics are bolded/underline}
      \centering
        \resizebox{\linewidth}{!}{%
        \begin{tabular}{ccccccc}
        \toprule[1.5pt]
            &&\model& Random Walk & Random Walk (*5) & Random Walk (*10) & Enumeration
            \\
            \hline
            \multirow{4}{*}{YAGO26K-906}&
             Hits@1 & \underline{0.146} & 0.066 & 0.114& 0.121& \textbf{0.312}
             \\&
             Hits@3 & \underline{0.239} & 0.097 & 0.169& 0.191& \textbf{0.452}
             \\&
             Hits@10 & \underline{0.376} & 0.139 & 0.220 & 0.288& \textbf{0.564}
             \\&
             MRR & \underline{0.218} & 0.089 & 0.151 & 0.174& \textbf{0.400}
             \\
            \hline
            \multirow{4}{*}{DB111K-174}&
            Hits@1 & \textbf{0.638} & 0.010 & 0.032& 0.070& -
             \\&
             Hits@3 & \textbf{0.825} & 0.014 & 0.048& 0.105& -
             \\&
            Hits@10 & \textbf{0.851} & 0.017 & 0.062 & 0.137& -
             \\&
             MRR & \textbf{0.731} & 0.012 & 0.043 & 0.093& -\\
             \bottomrule[1.5pt]
        \end{tabular}
        }
    
    \bigskip
    \resizebox{\linewidth}{!}{%
        \begin{tabular}{ccccccc}
        \toprule[1.5pt]
            &&\model& Random Walk & Random Walk (*5) & Random Walk (*10) & Enumeration
            \\
            \hline
            \multirow{3}{*}{YAGO26K-906}&
             \#Output Metapaths & 20,326 & 19,037 & 92,003& 187,400& 14,894,588
             \\&
             \#Valid Metapaths & 2,687 & 387 & 1,561& 2,940& 152,502
             \\&
             Valid Rate & \pmb{13.22\%} & 2.03\% & 1.70\% & 1.57\%& 1.02\%
             \\
            \hline
            \multirow{3}{*}{DB111K-174}&
            \#Output Metapaths & 30,847 & 22,506 & 111,892& 222,143& -
             \\&
             \#Valid Metapaths & 979 & 111 & 448& 788& -
             \\&
            Valid Rate & \pmb{3.17\%} & 0.49\% & 0.40\% & 0.35\%& -
             \\
             \bottomrule[1.5pt]
        \end{tabular}
    \label{table_multi_inductive}
        }
    % \begin{minipage}{.2\linewidth}
    %   \centering
%         \begin{tabular}{cccc}
%     \toprule[1.5pt]
%    & \#Output Metapaths & \#Valid Metapaths & Valid Rate 
% \\
%      \hline
%      \multicolumn{4}{c}{YAGO26K-906}
% \\
% \model&$20,326$&$2,687$&\pmb{$13.22\%$} 
% \\
% Random Walk&$19,037$&$387$&$2.03\%$ 
% \\
% Random Walk (*5)&$92,003$&$1,561$&$1.70\%$ 
% \\
% Random Walk (*10)&$187,400$&$2,940$&$1.57\%$ 
% \\
% Enumeration&$14,894,588$&$152,502$&$1.02\%$ 
% \\
% \hline
%      \multicolumn{4}{c}{DB111K-174}
% \\
% \model&$30,847$&$979$&\pmb{$3.17\%$}
% \\
% Random Walk&$22,506$&$111$&$0.49\%$ 
% \\
% Random Walk (*5)&$111,892$&$448$&$0.40\%$ 
% \\
% Random Walk (*10)&$222,143$&$788$&$0.35\%$ 
% \\
% Enumeration&-&-&-
% \\
% \bottomrule[1.5pt]
% \end{tabular}
\end{table*}

\begin{table}[tb]
    \centering
       \caption{Meta-path space for some relations in YAGO26K-906}
\resizebox{\linewidth}{!}{%
 \begin{tabular}{cccc}
    \toprule[1.5pt]
   & \#Possible Metapaths & \#Valid Metapaths & Valid Rate 
\\
\hline
wasBornIn&$4,890,928$&$59,520$&$1.22\%$ 
\\
LivesIn&$3,937,737$&$29,751$&$0.76\%$ 
\\
hasChild&$2,172,857$&$12,469$&$0.57\%$ 
\\
isCitizenOf&$1,810,611$&$39,778$&$2.20\%$ 
\\
PlaysFor&$1,075,622$&$1,143$&$0.14\%$ 
\\
isLeaderOf&$1,006,833$&$9,511$&$0.94\%$ 
\\
\bottomrule[1.5pt]
\end{tabular}
}
\label{table_metapath_space}
\end{table}

\subsection{Multi-relation Experiments} 
We evaluate the multi-relation transductive and inductive capabilities of \model~by query answering tasks\cite{das2018minerva}. Specifically, given the query $(e_h, r_q, ?)$, we aim to identify the tail entity $e_t$.
In multi-relation inductive experiments, the agent learns the meta-path generation mechanism for the trained relations and infers rewarding meta-paths for the test relations. 

\noindent \textbf{Datasets.} We use two large KBs: YAGO26K-906 and Dbpedia for multi-relation settings. The meta-path space for some relations in YAGO26K-906 is showcased in Table \ref{table_metapath_space}. We can see that, even in YAGO26K-906 which has comparatively fewer entity types and relations, each relation entails a million-scale meta-path space to search over and has extremely low valid rate. Here, a meta-path is valid if it has any meta-path instance in the $\mathcal{G}_I$, and low valid rate reflects the difficulty of mining meaningful meta-paths from the meta-path space. Besides, for DB111K-174, which contains 305 relations, we train and test on sampled relations. A detailed description on the two datasets and the chosen train/test relations could be found at Appendix \ref{appendix_dataset}. 

\noindent \textbf{Inference.} During inference, we utilize \model~to generate a set of high-quality meta-paths to reason about each test relation by employing a beam search with a large beam width of 400. 
Afterwards, the confidence of each outputted meta-path with regard to the queried relation $r_q$ is calculated, and is used as the score to infer tail entities.
Given $(e_h, r_q, ?)$, we apply the mined meta-paths on $e_h$ and rank possible tail entities after aggregating (max-pooling) the confidence of each meta-path.
Entities that are inaccessible from $e_h$ via any meta-path are given a rank of infinite value. 
We use the standard metrics for KB completion tasks, i.e., the Hits@1, 3, 10 and the mean reciprocal rank (MRR). All results are averaged over five independent runs.

% \begin{table*}[htb]
%     \centering
%     \renewcommand\arraystretch{1.13}
%        \caption{Transductive KG Query-answering results on YAGO26K-906 and DB111K-174, averaged over 5 runs. The best/second best metrics are bolded/underlined.}
% \resizebox{0.85\linewidth}{!}{%
%  \begin{tabular}{ccccccccc}
%     \toprule[1.5pt]
% &\multicolumn{4}{c}{YAGO26K-906}&\multicolumn{4}{c}{DB111K-174}
% \\
%    & Hits@1 & Hits@3 & Hits@10
%    & MRR & Hits@1 & Hits@3 & Hits@10
%    & MRR 
% \\
% \hline
% \model&0.157&0.250&\pmb{0.377}&0.223&0.668&\pmb{0.845}&\pmb{0.875}&\underline{0.754}
% \\
% TransE&0.092&0.225&0.316&0.176&0.538&0.809&0.858&0.675
% \\
% DistMult&0.039&0.052&0.118&0.059&0.359&0.481&0.555&0.431
% \\
% ComplEx&0.069&0.123&0.174&0.106&0.626&0.728&0.769&0.682 
% \\
% RotatE&\pmb{0.180}&\underline{0.260}&0.347&\pmb{0.237}&\pmb{0.768}&\underline{0.822}&0.849&\pmb{0.799}
% \\
% MINERVA&0.133&0.214&0.310&0.192&\underline{0.684}&0.786&\underline{0.863}&0.745 
% \\
% RNNLogic&\underline{0.161}&\pmb{0.263}&\underline{0.364}&\underline{0.225}&0.671&0.754&0.783&0.716
% \\
% \bottomrule[1.5pt]
% \end{tabular}
% }
% \label{table_multi_transductive}
% \end{table*}

\begin{table*}[h!]
    \centering
    \renewcommand\arraystretch{1.13}
       \caption{Transductive KG reasoning results on YAGO26K-906 and DB111K-174, averaged over 5 runs. The best/second best metrics are bolded/underlined.}
\resizebox{\linewidth}{!}{%
 \begin{tabular}{cccccccccc}
 \toprule[1.5pt]
 &&\model& TransE & DistMult & ComplEx & RotatE & MINERVA & RNNLogic & MLN4KB
 \\
 \hline
 \multirow{4}{*}{YAGO26K-906}& Hits@1 & 0.157 & 0.092 & 0.039 & 0.069& \pmb{0.180} & 0.133&\underline{0.161}
 & 0.128 \\
 & Hits@3 & 0.250 & 0.225 & 0.052 & 0.123& \underline{0.260} & 0.214&\pmb{0.263} & 0.236
 \\
 & Hits@10 & \pmb{0.377} & 0.316 & 0.118 & 0.174& 0.347 & 0.310&\underline{0.364} & 0.342
 \\
 & MRR & 0.223 & 0.176 & 0.059 & 0.106& \pmb{0.237} & 0.192&\underline{0.225} & 0.213
 \\
 \hline
 \multirow{4}{*}{DB111K-174}& Hits@1 & 0.668 & 0.538 & 0.359 & 0.626& \pmb{0.768} & \underline{0.684}&0.671 & 0.669
 \\
 & Hits@3 & \pmb{0.845} & 0.809 & 0.481 & 0.728& \underline{0.822} & 0.786&0.754 & 0.722
 \\
 & Hits@10 & \pmb{0.875} & 0.858 & 0.555 & 0.769& 0.849 & \underline{0.863}&0.783 & 0.785
 \\
 & MRR & \underline{0.754} & 0.675 & 0.431 & 0.682& \pmb{0.799} & 0.745&0.716 & 0.710
 \\
 \bottomrule[1.5pt]
\end{tabular}
}
\label{table_multi_transductive}
 \end{table*}

\subsubsection{Multi-relation Inductive Experiments} 
\label{sec:experiment:multi-relation_inductive}
\noindent \textbf{Baselines.} We notice that there is no existing method in the literature that could learn rules for relations without specific evidence. To provide a fair comparison, we test \model~against random walk and breadth-first search based enumeration in this setting. We set the same search attempts for \model~and random walk and we also evaluate the effect of increasing the search attempts for random walk (to 5 times and 10 times the normal attempts).

\noindent \textbf{Observations. }Table \ref{table_multi_inductive} displays the multi-relation inductive results on YAGO26K-906 and DB111K-174. \model~significantly outperforms random walk in terms of efficiency and effectiveness. Even when we increase the search attempts for random walk by a factor of ten, it still cannot beat the accuracy of our approach on YAGO26K-906. On DB111K-174, \model~exhibits more significant superiority over the random walk method.
% On DB111K-174, the superiority of \model~over random walk is more significant.
Enumeration presents the best prediction answers on YAGO26K-906, but at an enormous computational cost (consuming nearly 498 times the inference time than \model). Furthermore, enumeration cannot scale to DB111K-174 given the enormous meta-path space.
Notably, \model~produces meta-paths with significantly higher valid rate compared to other methods, which is about 6.5 times the valid rate obtained using random walk, and 13 times obtained using enumeration. This indicates that \model~indeed learns the mechanism to generate meaningful and relevant meta-paths.
\subsubsection{Multi-relation Transductive Experiments}
\label{sec:experiment:multi-relation_transductive}
\noindent \textbf{Baselines.} We compare \model~with six widely-used KB reasoning baselines: TransE~\cite{bordes2013transe}, DistMult~\cite{yang2014Distmult}, ComplEx~\cite{trouillon2016complex}, RotatE~\cite{sun2019rotate}, MINERVA~\cite{das2018minerva}, RNNLogic~\cite{qu2020rnnlogic} and MLN4KB~\cite{fang2023mln4kb}. RNNLogic could hardly scale to DB111K-174 and we restrict the rule length to 3.
Detailed information on their implementation could be referenced in Appendix~\ref{appendix_baseline}.

\noindent \textbf{Observations. }In Table \ref{table_multi_transductive}, we show the multi-relation transductive results on YAGO26K-906 and DB111K-174. \model~demonstrates superior performance over three embedding-based methods, namely TransE, DistMult, and ComplEx, and wins a noticeable advantage over the path-based MINERVA and the rule-based MLN4KB. On YAGO26K-906, \model~outperforms all other models in the Hits@10 metric and performs comparably with the rule-learning RNNLogic across all metrics. Additionally, on DB111K-174, \model~achieves the highest scores for Hits@3, Hits@10, and the second-highest score for MRR. Overall, \model~competes strongly with state-of-the-art models for transductive KG query-answering.

Although RotatE exhibits fair competitiveness with \model~on both datasets, it is prone to being influenced by unseen entities due to its embedding-based nature. We would further elaborate this issue in Section \ref{sec:inductive}. Besides, RNNLogic encounters scalability issues and takes a remarkably long time of 7 days to complete even one run on DB111K-174.

\subsubsection{Comparing the Multi-relation Inductive and Multi-relation Transductive Capacity of \model}
\begin{figure*}[htb]
\centering
\includegraphics[width=\linewidth]{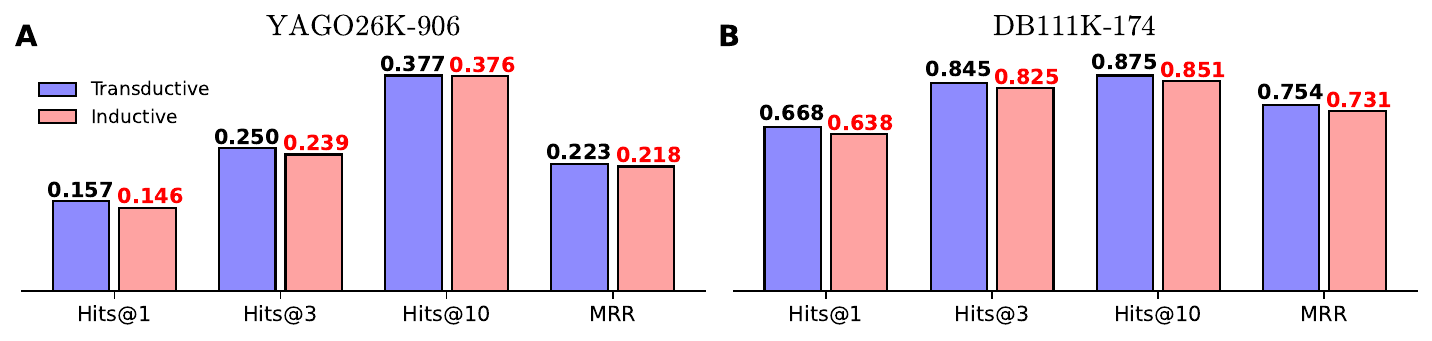}
\caption{
% The difference between \model's performances in multi-relation transductive experiments and multi-relation inductive experiments
The performance difference of \model~between multi-relation transductive setting and multi-relation inductive setting.
}\label{fig_diff}
\end{figure*} 

Fig. \ref{fig_diff} shows the performance drop of \model~from multi-relation inductive experiments to multi-relation transductive ones. Impressively, the performance degradation is negligible across all metrics, with an average decline rate of merely $3.32\%$. This observation suggests that \model~can effectively learn a policy to establish high-quality meta-paths by examining meta-path samples (and their scores) generated during training on some relations. With this policy, the model can efficiently predict meta-paths for test relations by referring to the trained schema-level representations.

\subsection{Per-relation Experiments} 
\label{sec:experiment:per-relation_transductive}
% We demonstrate the effectiveness of~\model~in reasoning schema-complex and general HIN following the link prediction evaluation practices in prior works~\cite{meng2015fspg,wan2020mpdrl}.
In the HIN literature~\cite{meng2015fspg,wan2020mpdrl}, the effectiveness of models is typically evaluated in a transductive manner and relation-wise. Following this established practice, we demonstrate the effectiveness of~\model~in reasoning over schema-complex and general HINs in this section.

\noindent \textbf{Baselines.}
%We compare \model~with eight existing relation reasoning methods falling in the classes of meta-path-based, embedding-based and multi-hop methods, as in Table \ref{tab:baseline}.
We compare \model~with eight existing relation reasoning methods: MPDRL~\cite{wan2020mpdrl}, PCRW~\cite{lao2010pcrw}, Autopath~\cite{yang2018autopath}, Metapath2Vec~\cite{dong2017metapath2vec}, HIN2Vec~\cite{fu2017hin2vec}, NSHE~\cite{zhao2020network}, RotatE~\cite{sun2019rotate},  TransE~\cite{bordes2013transe}, MINERVA~\cite{das2018minerva}. 
See Appendix~\ref{appendix_baseline} for their detailed information.

\noindent \textbf{Datasets.}
To validate \model’s relation-wise reasoning ability on schema-complex HINs, we run experiments on YAGO26K-906 and NELL.
% , which provide both instance and schema-level information.
% Although there are other large-scale KBs, most of them do not provide schema-level information.
We also include a schema-simple HIN, Chem2Bio2RDF, to demonstrate that \model\ is also useful for general HINs. The details about the chosen datasets are in Appendix \ref{appendix_dataset}. We consider three relations each for the two KBs: \{\textit{isCitizenOf}, \textit{DiedIn}, \textit{GraduatedFrom}\} for YAGO26K-906 and \{\textit{WorksFor}, \textit{CompetesWith}, \textit{PlaysAgainst}\} for NELL. For Chem2Bio2RDF, we focus on predicting drug-target connectivity, i.e., the relation \textit{bind}.

\begin{table*}[htb]
    \centering
    \renewcommand\arraystretch{1.13}
       \caption{Per-relation link prediction results for YAGO26K-906, NELL and Chem2Bio2RDF, averaged over 5 runs. The bold/underlined results respectively indicate the best/second best performances for each relation among all methods.}
\resizebox{1\linewidth}{!}{%
 \begin{tabular}{cc>
 {\centering}m{0.08\textwidth}>{\centering}m{0.05\textwidth}>
 {\centering}m{0.05\textwidth}>{\centering}m{0.05\textwidth}>{\centering}m{0.08\textwidth}>{\centering}m{0.06\textwidth}>{\centering}m{0.04\textwidth}>{\centering}m{0.05\textwidth}>{\centering}m{0.05\textwidth}c}
    \toprule[1.5pt]
   &&\model & MPDRL & PCRW
   & Autopath & Metapath2Vec & HIN2Vec & NSHE & RotatE & TransE & MINERVA 
   \\
     \hline
     \multicolumn{11}{c}{YAGO26K-906}
     \\
      \multirow{2}{*}{isCitizenOf}&ROC-AUC&$\pmb{0.865}$&$0.778$&$0.584$&$0.755$&$0.652$&$0.800$&0.788 &$0.778$&$0.590$& \underline{$0.828$} 
       \\
      &AP&$\pmb{0.904}$&$0.793$&$0.706$&$0.723$&$0.781$&$0.837$&0.803 &$0.830$&$0.810$& \underline{$0.840$}
      \\
           
      \multirow{2}{*}{DiedIn}&ROC-AUC&$\pmb{0.947}$&$0.710$&$0.645$&$0.723$&$0.661$&$0.785$&0.748 &\underline{$0.860$}&$0.622$&$0.632$
       \\
       &AP&$\pmb{0.971}$&$0.679$&$0.686$&$0.787$&$0.830$&$0.877$&0.778 &\underline{$0.907$}&$0.745$&$0.786$
      \\
                
      \multirow{2}{*}{GraduatedFrom}&ROC-AUC&$\pmb{0.896}$&$0.664$&$0.586$&$0.724$&$0.661$&$0.803$&0.815 &\underline{$0.830$}&$0.662$&$0.609$
       \\
       &AP&$\pmb{0.912}$&$0.743$&$0.664$&$0.718$&$0.783$&$0.842$&0.839 &\underline{$0.855$}&$0.750$&$0.718$
      \\
      \hline
     \multicolumn{11}{c}{NELL}
     \\
      \multirow{2}{*}{WorksFor}&ROC-AUC&$\pmb{0.913}$&$0.759$&$0.646$&$0.703$&$0.613$&$0.790$&0.774 &\underline{$0.868$}&$0.637$&$0.767$
       \\
       &AP&$\pmb{0.945}$&$0.871$&$0.735$&$0.778$&$0.819$&$0.870$&0.859 &\underline{$0.911$}&$0.719$&$0.802$
      \\
                   
      \multirow{2}{*}{PlaysAgainst}&ROC-AUC&\underline{$0.907$}&$0.774$&$0.567$&$0.544$&$0.761$&$0.783$&0.814 &$\pmb{0.966}$&$0.845$&$0.541$
       \\
       &AP&\underline{$0.938$}&$0.823$&$0.745$&$0.691$&$0.904$&$0.917$&0.883 &$\pmb{0.993}$&$0.928$&$0.683$
      \\
                   
      \multirow{2}{*}{CompetesWith}&ROC-AUC&$0.706$&$0.589$&$0.547$&$0.567$&$0.600$&$0.730$&0.677 &$\pmb{0.870}$&\underline{${0.774}$}&$0.585$
       \\
       &AP&$0.806$&$0.695$&$0.699$&$0.685$&$0.735$&$0.867$&0.784 &$\pmb{0.939}$&\underline{$0.908$}&$0.702$
      \\
        \hline
     \multicolumn{11}{c}{Chem2Bio2RDF}
     \\
       \multirow{2}{*}{Bind}&ROC-AUC&$\pmb{1.000}$&$0.813$&$0.609$&$0.685$&$0.759$&$0.880$&0.901 &\underline{$0.936$}&$0.909$&$0.715$
       \\
       &AP&$\pmb{1.000}$&$0.890$&$0.723$&$0.792$&$0.914$&$0.939$&0.963 &\underline{$0.984$}&$0.978$&$0.816$
      \\
       \bottomrule[1.5pt]
   \end{tabular}
   }
   \label{table_link_result}
\end{table*}

\noindent \textbf{Dataset Preparation.} To predict relation $r_q$ with meta-paths of maximum length $l$, we list all $r_q$-connected entity pairs and check each pair using a breadth-first search to determine whether there exists an instance path connecting them within $l-1$ hops (apart from directly via $r_q$).
The entity pairs that fail this check are filtered out.
Unlike the practice in MPDRL\cite{wan2020mpdrl}, we do not set maximum number of attempts for the search to help include more condition-satisfying entity pairs for fair comparison.
% connected by long paths 
The remaining pairs form the dataset, which is split into train/test set in an $8:2$ ratio. Before training, facts in the test set are removed from the instance graph, providing the groundings to calculate coverage and confidence.

\noindent \textbf{Training Settings for \model.} On the schema graph, a relation could connect multiple entity type pairs $(t_{src}, t_{tgt})$. On each attempt, the agent chooses one such pair to form a query, starts at $t_{src}$ and finds a rewarding path to $t_{tgt}$. For YAGO26K-906 and Chem2Bio2RDF, the query set for training is all possible entity type pairs.
%the training objective is simply all possible entity type pairs connected by $r_q$.
However, in NELL, some relations could connect an enormous number of entity type pairs (e.g., 1488 for \textit{CompetesWith}) and many of these type pairs map to very few entity pairs. 
%Many useful meta-paths would and if we egalitarianly mine meta-paths for these type pairs, which are not equally useful, we would fail to discover some meta-paths that are applied to a great deal of instance pairs. 
%Therefore, we narrow the training objective on NELL by checking all instance pairs related with $r_q$ in the training set and focus on finding meta-paths for the least number of type pairs that cover a threshold percent (set as $80\%$) of the instance pairs. 
Therefore, for \model, we narrow the query set on NELL by the minimum type pairs that cover a threshold percent (set as $80\%$) of the $r_q$-related entity pairs. See Appendix \ref{appendix_hyper} for further training settings.

\noindent \textbf{Link Prediction Settings.} For link prediction, the test set mentioned above constitutes the positive pairs. Following the practice in~\cite{wan2020mpdrl}, we generate the negative pairs by replacing the target entity in instance graph samples with a fake one but of the same type. All negative samples for the six relations in YAGO26K-906 and NELL are generated in this manner. The Chem2Bio2RDF already includes such negative dataset and we adopt it directly. The positive/negative rate is 2:1. To calculate similarity for each pair, we sum up the confidence of all the meta-paths that connect the pair. Ultimately, we adopt a linear regression model with L1-regularization to perform link prediction. For the four embedding-based approaches, we calculate the Hadamard product of the head and tail entity embeddings and use an SVM classifier for link prediction.

\noindent \textbf{Evaluation Metrics.}
We use two diagnostic metrics: the area under the receiver operating characteristic curve (ROC-AUC) and average precision (AP). 
The results for each method were averaged over five independent runs.

\begin{figure*}[htb]
\centering
\includegraphics[width=0.95\linewidth]{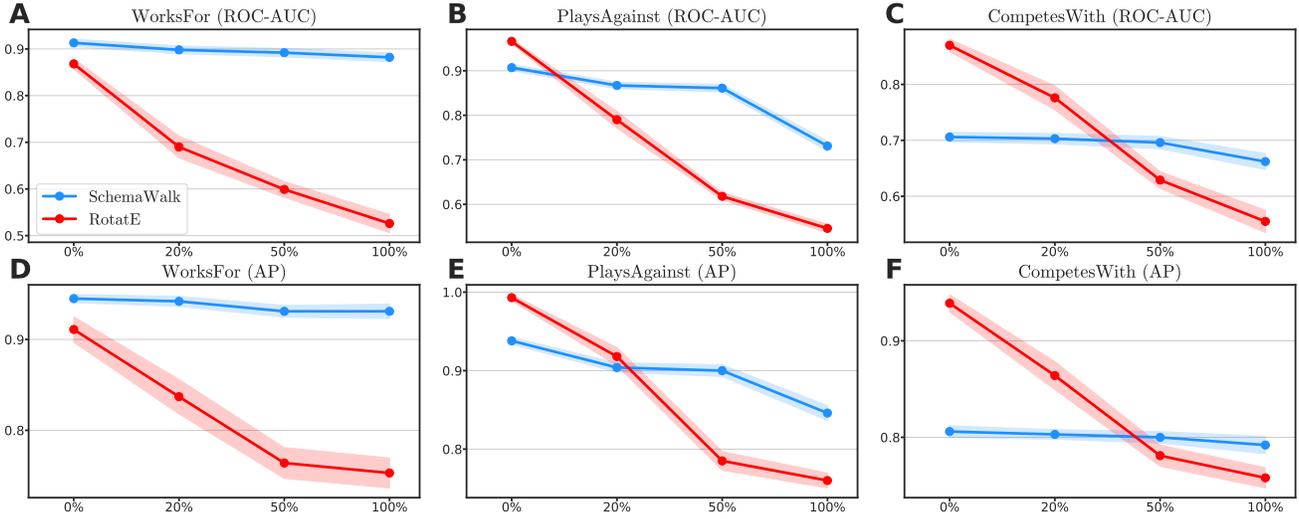}
\caption{Entity-level inductive experiment results for \model~(Blue) and RotatE (Red). The above/below three charts respectively show the ROC-AUC/AP values. The horizontal axes represent the node removal rate. The shadow area represents the confidence intervals over 5 runs.}\label{fig_inductive}
\end{figure*}

\subsubsection{Results on Schema-complex HINs} \label{sec:result_syn}
The ROC-AUC and AP for \model~and the baselines for the link prediction task on the six selected relations are presented in Table \ref{table_link_result}. \model~excels in all relations in YAGO26K-906 and the relation \textit{WorksFor} in NELL, only lagging behind by RotatE for relations \textit{PlaysAgainst} and \textit{CompetesWith} in NELL. 

Among the meta-path based baselines, HIN2Vec achieves decent prediction accuracy. It generates superior embeddings compared to Metapath2Vec, by utilizing meta-paths for multiple relations.
The performance of NSHE is also reasonably good, similar to HIN2Vec, due to the high-order structure-derived embeddings it obtains.
% and is less susceptible to performance drop when the schema graph gets more complex.
MPDRL produces relatively satisfactory results, thanks to its RL-based strategy to explore instance paths. Autopath exhibits acceptable capability, but its ability is confined by the limited number of discovered meta-paths, particularly for relations that involve an extensive meta-path space. PCRW is the most mediocre model among them because of its randomness-based nature. For the multi-hop-based MINERVA, it performs well in reasoning \textit{isCitizenOf} and \textit{WorksFor}, but performs poorly for other relations. We observe that embedding-based approach for KBs, particularly RotatE, dominates the two relations in NELL. Upon further inspection, we notice that both \textit{PlaysAgainst} and \textit{CompetesWith} involve thousands of type pair, suggesting an enormous meta-path space (ten million-scale) to explore. 

\begin{figure*}[htb]
\centering
\includegraphics[width=0.95\linewidth]{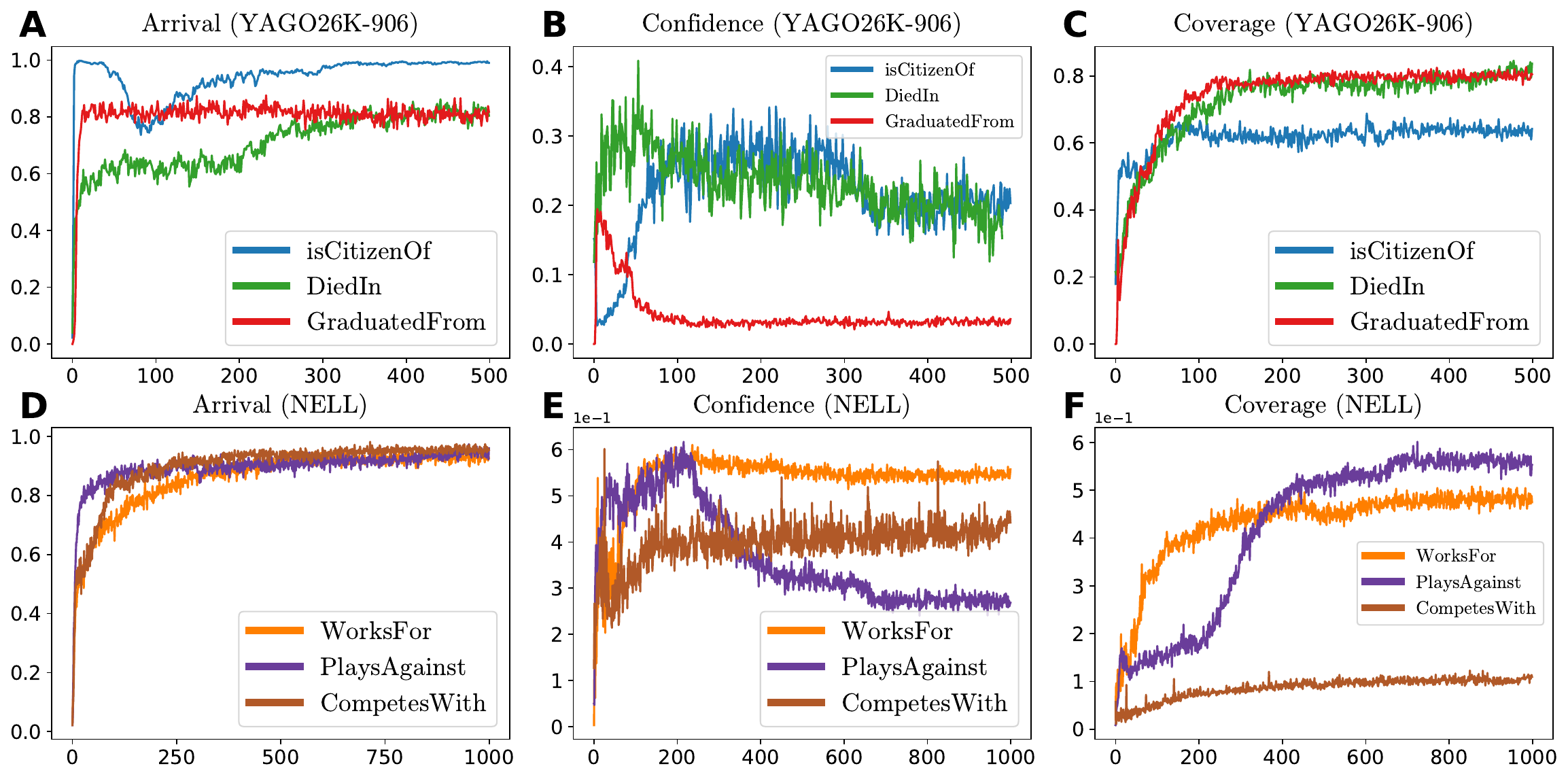}
\caption{\model's training curves for arrival rate, confidence and coverage on YAGO26K-906 and NELL, averaged over 5 runs.}\label{fig_train_converge}
\end{figure*} 

\subsubsection{Results on Schema-simple HIN}
We show the link prediction results on Chem2Bio2RDF in Table \ref{table_link_result}. As demonstrated by the results, the meta-paths generated by \model~can distinguish all positive facts from negative facts. Besides, \model~only takes an average of 30 iterations (or roughly 400 seconds) to learn all the necessary meta-paths for link prediction. In contrast, other baselines are much more time-consuming, often taking several hours due to the gigantic instance graph of Chem2Bio2RDF. Our model is effective and efficient for general HINs.

\subsubsection{Entity-level Inductive Experiment}\label{sec:inductive}
In the Entity-level inductive experiment, we concentrate on the NELL dataset and compared our method against RotatE. We randomly sample $40\%$ of the positive testing set, remove $0\%$, $20\%$, $50\%$ and $100\%$ of the nodes that appear in the pairs from the instance graph, followed by regular training and link prediction. The ROC-AUC and AP results for the two methods under the four scenarios are shown in Fig. \ref{fig_inductive}.

We discover that, with the removal rate of only $20\%$, \model~already beats RotatE for relation \textit{PlaysAgainst} in terms of ROC-AUC and wins an all-out victory for all selected relations with the removal rate over $50\%$. The performance of RotatE deteriorates drastically with the removal rate, whilst \model~only projects minor signs of being affected. 
We only observe relatively evident performance drop for \model~when reasoning \textit{PlaysAgainst}, as the removal rate reaches $100\%$ from $50\%$ (ROC-AUC dropped by $15.1\%$ and AP dropped by $6\%$).
We attribute this phenomenon to the removal of some key instance nodes, which cut off massive instance paths and drastically affect the coverage and confidence score.

\begin{table*}[htb]
\centering
\caption{Example meta-paths outputted/inferred by \model\ %with their coverage and confidence
}\label{tab_example_metapath}
\resizebox{0.85\linewidth}{!}{%
\begin{tabular}{c|c|c|c}
\toprule[1.5pt]
\textbf{Relations}& \multicolumn{1}{c|}{\textbf{Meta-path}}&\textbf{Cvrg.}&\textbf{Conf.}
\\ \midrule[1.1pt]
\multirow{5}{*}{WorksFor (NELL)}   
&$\mbox{Person} \xrightarrow{BelongsTo}\mbox{Company}$      &  0.505 & 0.818                   \\
& $\mbox{Person} \xrightarrow{CollaboratesWith}\mbox{Company}$      &  0.564 & 0.460                   \\
& $\mbox{Person} \xrightarrow{Controls}\mbox{Company}$      &  0.152 & 0.767                   \\
&$\mbox{CEO} \xrightarrow{Leads}\mbox{Company}$      &  0.882 & 0.745                   \\
& $\mbox{Journalist} \xrightarrow{WritesFor}\mbox{Newspaper}$      &  0.741 & 0.399                   \\
\midrule[1.1pt]
\multirow{5}{*}{PlayAgainst (NELL)}   
&$\mbox{Team} \xrightarrow{CompetesWith}\mbox{Team}$      &  0.149 & 0.975                   \\
& $\mbox{Team} \xrightarrow{PlaysIn}\mbox{League}\xrightarrow{SuperPartOf}\mbox{Team}$      &  0.525 & 0.295                   \\
& $\mbox{Team} \xrightarrow{WinTrophy}\mbox{Game}\xrightarrow{Participant}\mbox{Team}$      &  0.018 & 0.782                   \\
& $\mbox{Team}\xrightarrow{KnownAs}\mbox{Team}\xrightarrow{CompetesWith}\mbox{Team}$      &  0.305 & 0.579                   \\

&$\mbox{Coach} \xrightarrow{BelongsTo}\mbox{League}\xrightarrow{LeagueTeam}\mbox{Team}$      &  0.707 & 0.212                   \\
\midrule[1.1pt]
\multirow{5}{*}{isCitizenOf (YAGO26K-906)}   
& $ \mbox{Person} \xrightarrow{BornIn}\mbox{District}\xrightarrow{LocatedIn}\mbox{Country}$      &  0.295 & 0.083                   \\
& $\mbox{Person} \xrightarrow{DiedIn}\mbox{District}\xrightarrow{LocatedIn}\mbox{Country}$      &  0.148 & 0.070                   \\
&  $\mbox{Person} \xrightarrow{LivesIn}\mbox{Country}$      &  0.082 & 0.152                   \\
&  $\mbox{Scientist} \xrightarrow{GraduatedFrom}\mbox{University}\xrightarrow{LocatedIn}\mbox{Country}$      &  0.167 & 0.540                   \\
&  $\mbox{Scientist} \xrightarrow{WorksAt}\mbox{University}\xrightarrow{isLocatedIn}\mbox{Country}$      &  0.205 & 0.421                  \\
\midrule[1.1pt]
\multirow{5}{*}{GraduatedFrom (YAGO26K-906)}   
& $\mbox{Person} \xrightarrow{isCitizenOf}\mbox{Country}\xrightarrow{LocatedIn^{-1}}\mbox{University}$      &  0.169 & 0.034                   \\
& $\mbox{Person} \xrightarrow{AdvisedBy}\mbox{Scientist}\xrightarrow{WorksAt}\mbox{University}$      &  0.080 & 0.260                   \\
&$\mbox{Politician}\xrightarrow{isPoliticianOf}\mbox{Country}\xrightarrow{LocatedIn^{-1}}\mbox{University}$      &  1.000 & 0.133                   \\
&$\mbox{Scientist} \xrightarrow{WorksAt}\mbox{University}$      &  0.268 & 0.191                   \\
& $\mbox{Scientist} \xrightarrow{MarriedTo}\mbox{Scientist}\xrightarrow{WorksAt}\mbox{University}$      &  0.030 & 0.125                   \\

\midrule[1.1pt]
\multirow{5}{*}{StateOfOrigin (DB111K-174, \textbf{Inferred})}   
&$\mbox{Person}\xrightarrow{Nationality}\mbox{Country}$      &  0.651 & 0.980                   \\
& $\mbox{Person}\xrightarrow{BirthPlace}\mbox{Country}$      &  0.261 & 0.087                   \\
&$\mbox{Person}\xrightarrow{DeathPlace}\mbox{Country}$      &  0.091 & 0.080                   \\
&
$\mbox{Person} \xrightarrow{Residence}\mbox{Country}$      &  0.063 & 0.204                   \\

& $\mbox{Person} \xrightarrow{Predecessor}\mbox{Person}\xrightarrow{Nationality}\mbox{Country}$      &  0.011 & 0.892                   \\

\midrule[1.1pt]
\multirow{5}{*}{MusicalBand (DB111K-174, \textbf{Inferred})}   

&$\mbox{Single}\xrightarrow{byMusicalArtist}\mbox{Band}$      &  0.444 & 1.000                   \\
& $\mbox{Single}\xrightarrow{Producer}\mbox{Band}$      &  0.107 & 0.765                   \\
&$\mbox{Single}\xrightarrow{Writer}\mbox{Band}$      &  0.140 & 0.129                   \\
&
$\mbox{Single} \xrightarrow{Producer}\mbox{MusicalArtist}\xrightarrow{associatedBand}\mbox{Band}$      &  0.135 & 0.122                   \\

& $\mbox{Single} \xrightarrow{subsequentWork}\mbox{Single}\xrightarrow{byMusicalArtist}\mbox{Band}$      &  0.126 & 0.719                   \\
\bottomrule[1.5pt]
\end{tabular}}
\end{table*}

\section{Further Analysis}
\subsection{Convergence Properties}\label{appendix_training_convergence}
We present the training curves for arrival rate, confidence and coverage on relations in YAGO26K-906 and NELL, as displayed in Fig. \ref{fig_train_converge}. 
All curves exhibit an upward trend as the training starts. Most of the curves for arrival rate and coverage continue to grow until convergence. Conversely, the confidence curves for some relations initially ascend to a high point, but converge to a sub-optimal value eventually. This happens because the agent prioritizes coverage over confidence at times, believing that finding a set of rules that covers the majority of entity pairs is more crucial than identifying only a few rules with the highest confidence.

During the initial training period, \model~exhibits noticeable exploratory behaviour, thanks to the regularization item. Nevertheless, this behaviour does not adversely affect the learning of the agent, as the confidence and coverage curves converge eventually, even for relations \textit{PlaysAgainst} and \textit{CompetesWith} that connect thousands of entity type pairs. The arrival rate curves tend to converge faster compared to the confidence and coverage curves, indicating that the agent needs to know how to navigate to the target type node before learning to generate high quality meta-paths. 
The majority of the confidence/coverage curves end up converging at different values, reflecting the variations in the confidence/coverage distribution for meta-paths amongst different relations. 
%indicating that the agent has formed meta-paths before it learns to generate meta-paths with high coverage rate. 
Lastly, all curves converge before the set number of iterations and further training iterations do not empirically yield better prediction results.

With aforementioned notations, the overall complexity for the training process is bounded by $\mathcal{O}(KN(l-1+(|V|\Delta(\mathcal{G}_I)^2)^{l-2})$ per epoch (See Appendix~\ref{appendix_time_train_complex} for detailed analysis). Even when training on the most complex schema graph, that of NELL, our method demonstrates remarkable efficiency. The observed wall times for our model to complete a single epoch for the relations \textit{PlaysAgainst}, \textit{CompetesWith}, and \textit{PlaysFor} are 1.44s, 1.98s, and 5.01s respectively. Consequently, it took only about 868s, 793s, and 1353s respectively to converge to a reasonable point for these relations. This observation affirms that our strategy, which involves searching for the target type node within the schema graph, delivers efficient training.

% For \textit{PlaysAgainst}, \model\ also exhibits noticeable exploratory behaviour to find meta-paths with high coverage before shifting to maximise confidence.
%the highest confidence value of the meta-paths for the query relation\FC{It is unclear to me}. 
% We also see that, for relations \textit{DiedIn} and \textit{CompetesWith}, the confidence/coverage curves still have the potential to reach a higher point. However, we do not observe better link prediction results with more training iterations for the time being.

\subsection{Meta-paths Analysis}
\label{appendix_metapaths_dis}
We examine the generated meta-paths and select those meta-paths with high rewards, which are presented in Table \ref{tab_example_metapath}. 
A careful observation of these meta-paths reveals their significant diversity and contextual relevance. For instance, our model can learn multiple meta-paths to deduce citizenship via birthplace, alma mater, or workplace. Furthermore, it learns meta-paths not just for top-hierarchy types like \textit{Person}, but also for specific types from lower hierarchies like \textit{CEO} and \textit{Scientist}. In terms of semantic correctness, each of these meta-paths precisely encapsulates and expresses the semantic meaning of the corresponding relations. 
An example includes our model learning the meta-path that infers an individual's alma mater by taking into account the employment location of their university instructor.
% We can see that these meta-paths are meaningful and diverse.
% We could learn both general and specific meta-paths to deduce a relation.
We also observe that our model is capable of using synonymous relations to explain target relations, as demonstrated by utilizing \textit{CompetesWith} for \textit{PlayAgainst} and \textit{Nationality} for \textit{StateOfOrigin}).
In multi-relation inductive settings, \model~also infer high-quality meta-paths for untrained relations (e.g., \textit{StateOfOrigin} and \textit{MusicalBand} in DB111K-174). 

The coverage and confidence of above meta-paths are evaluated under the relevant KB. We thus observe that 16.7\% of the scientists entailing \textit{isCitizenOf} would have information of where their alma maters locate and, 54.0\% of the target countries indicated by such meta-path are the true answers.
% 13.4\% of the scientists satisfying such meta-path have relation \textit{isCitizenOf}.
%, which seems intuitive as many students study overseas\FC{Unclear to me.}. 
% It may seems counter-intuitive that the meta-path $ Person \xrightarrow{BornIn}District\xrightarrow{LocatedIn}Country$ or $Person \xrightarrow{DiedIn}District\xrightarrow{LocatedIn}Country$ has a high coverage rate but with a fairly low confidence. We remark that this is because of the incompleteness of KBs. 
% As missing links are prevalent in the KB world, the numerator in Eq.~(\ref{equ_conf}) is smaller than the counterpart
% %\FC{What is the counterpart}
% in the real world we perceive. Similarly, the meta-path $Person \xrightarrow{LivesIn}Country$ has a low coverage rate as there are only a few \textit{LivesIn} links annotated in the KB. 
We could also observe some interesting facts for politicians, scientists and coaches. For example, we find that 12.5\% of the scientists would graduate from the same university where their scientist spouse works.

\subsection{Parameter Analysis on $\lambda_1$ and $\lambda_2$}\label{further_lambda}
\begin{figure}[tb]
\centering
\includegraphics[width=\linewidth]{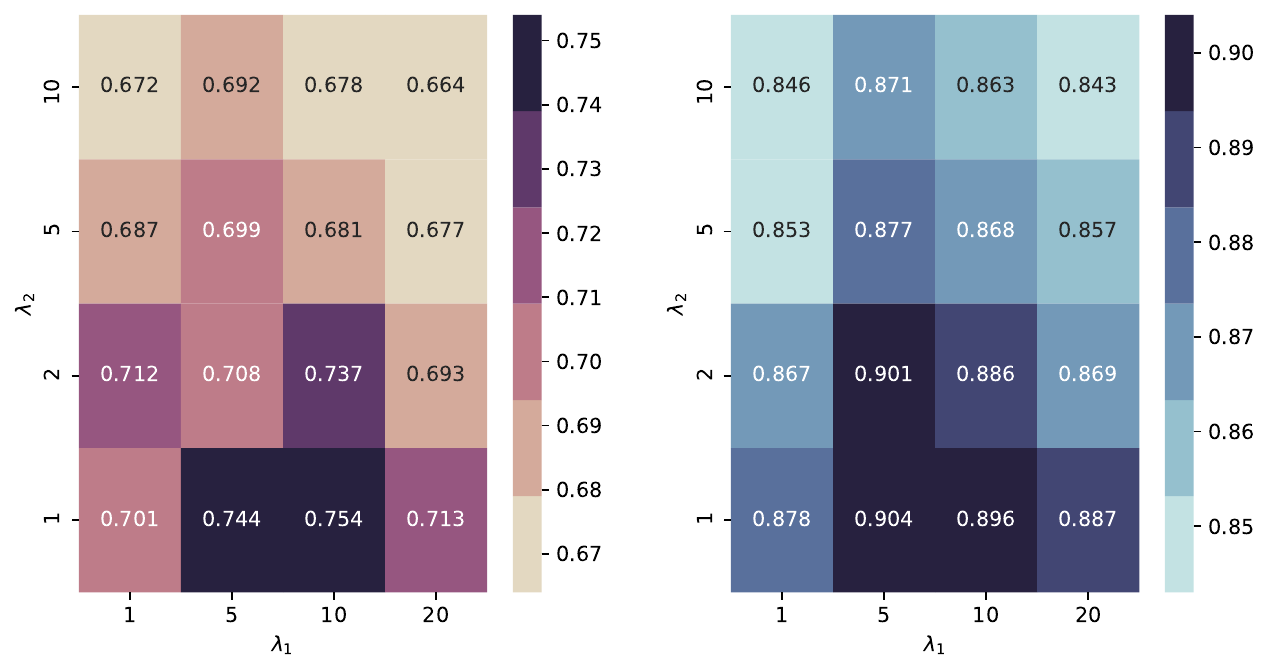}
\caption{The left and right plots respectively depict the results for transductive KG reasoning on DB111K-174 (measured by MRR) and per-relation link prediction on YAGO26K-906 with the relation \textit{isCitizenOf} (expressed in AUC) with varying values for $(\lambda_1, \lambda_2)$.}\label{fig_train_converge}
\end{figure} 
We scrutinize the impact of parameters $(\lambda_1, \lambda_2)$, which signify respective weights for coverage and confidence as expressed in Equation (\ref{equ_reward}), on the performance of our method. We selected values for $\lambda_1$ from the set $\{1,5,10,20\}$ and for $\lambda_2$ from $\{1,2,5,10\}$. The scale for these parameters is not identical due to a typically smaller value of coverage in relation to confidence.
With these selected parameters, we conducted a series of experiments for transductive KG reasoning on DB111K-174 and per-relation link prediction on the YAGO26K-906 (on relation \textit{isCitizenOf}). Our empirical observations indicate that the optimal values for $(\lambda_1, \lambda_2)$ for these distinct tasks are $(10, 1)$ and $(5, 1)$, respectively. 
Moreover, it is noteworthy that performance deteriorates by 12\% and 7\% respectively when $\lambda_1$ is excessively large. This downturn in performance is possibly due to the arrival item, devised to facilitate training, being swallowed up in the reward signal when in these extreme cases.
However, interestingly, even when faced with overly large values for $(\lambda_1, \lambda_2)$, the degradation in performance was not significant. This, in turn, underscores that our approach maintains a smooth performance trajectory irrespective of the values of $(\lambda_1, \lambda_2)$, denoting a low sensitivity to these parameters.

\subsection{Inference Time}\label{further_complexity}
We pose that our model offers superior efficiency during inference since it mainly search for the target entity type within its corresponding local neighborhood indicated by the schema graph. This sharply contrasts with the markedly lower efficiency of other meta-path discovery methods, which necessitate exhaustive exploration of the instance graph.
The primary computational expense during inference for~\model~rests in the computation of probabilities for all potential outgoing edges along the defined meta-path. Therefore, the inference time of~\model~is fundamentally contingent on the degree distribution of the graph.
Assuming the schema graph $\mathcal{G}_s$ conforms to a power law degree distribution, the test-time complexity per epoch, based on the paper's notations, is given as $\mathcal{O}(B'(l-1) \frac{\eta}{\eta-1})$, represents the power law's coefficient. 
Please refer to Appendix~\ref{appendix_infer_time_complexity} for an in-depth analysis.

% Assuming a power law degree distribution for the schema graph, a pattern seen frequently in natural graphs, we can deduce that the average inference time for our method approximates $O(\frac{\eta}{\eta-1})$, provided that the coefficient of the power law $\eta>1$. The median inference time for~\model~remains $O(1)$ for all values of $\eta$.}
% \input{Section/ablation}
\section{Conclusion and Future Work}
\label{conclusion}
In this paper, we investigate how to learn meta-paths in schema-complex HINs. 
% We successfully design an RL agent \model~that walks on the network schema, and learns from the rewards defined on the instance graph.
We successfully design a meta-path discovery framework, \model, which allows an RL agent to walk on the schema graph, and learn from the rewards defined on the instance graph. \model~offers several advantages. 
Firstly, among all meta-path reasoning methods, \model~is the pioneering multi-relation reasoning method for HINs, and can fascinatingly, predict fruitful meta-paths for relations without specific evidence.
Besides, in contrast to methods of learning meta-paths on the instance graph, \model\ is more effective, as it directly obtains high-quality meta-paths without requiring summarization from partial observations of path instances.
Additionally, in comparison to hand-crafted or graph traversal-based methods, \model\ is more efficient at exploring large combinatorial search space and can thus handle extensive schema graphs.
Finally, when compared to embedding-based approaches,~\model\ demonstrates superior instance-level explainability and generalization ability, particularly when reasoning about unseen entities, as it leverages the mined meta-paths rather than embedding vectors for predictions.

We present the first inductive meta-path learning model, which could be applied to reason multiple relations after a single training. The trained agent could not only learn high-quality meta-paths for a set of given relations, and also output rewarding meta-paths for untrained relations. The query answering and link prediction experiments fully verify the effectiveness of learning meta-paths in this way. The presented results also highlight the importance of meta-paths, not only in schema-simple HINs like Chem2Bio2RDF, but also in schema-complex HINs like KBs, where embedding-based models take dominant roles. Once high-quality meta-paths are obtained for schema-complex HINs, we could achieve better and explainable performances.

Future work includes at least four aspects. First, we could combine~\model's rewards with the evaluation metrics for downstream tasks and obtain more task-specific meta-paths. Besides, advances in graph representation learning and RL may enable us to design better agent and further improve the performance. Furthermore, we could build a score estimator to reduce the time to evaluate each meta-path during learning. Lastly, we could integrate temporal logic into our reasoning method, thereby making it applicable to temporal HIN.

\section*{Acknowledgments}
This work was supported in part by the National Natural Science Foundation of China (NSFC) under Grant 62206303 and Grant 62001495 and in part by China Postdoctoral Science Foundation under Grant 48919 and Science and Technology Innovation Program of Hunan Province (Grant Number: 2023RC3009). We would like to express our gratitude to Ziniu Hu for his valuable advice. We would also like to thank Jie Kang and Yuehang Si for providing us with some baseline results.

% {\appendix[Proof of the Zonklar Equations]
% Use $\backslash${\tt{appendix}} if you have a single appendix:
% Do not use $\backslash${\tt{section}} anymore after $\backslash${\tt{appendix}}, only $\backslash${\tt{section*}}.
% If you have multiple appendixes use $\backslash${\tt{appendices}} then use $\backslash${\tt{section}} to start each appendix.
% You must declare a $\backslash${\tt{section}} before using any $\backslash${\tt{subsection}} or using $\backslash${\tt{label}} ($\backslash${\tt{appendices}} by itself
%  starts a section numbered zero.)}

\bibliographystyle{IEEEtran}
\bibliography{ref}

 

% \end{IEEEbiography}

% \vspace{-33pt}
\begin{IEEEbiography}[{\includegraphics[width=1in,height=1.25in,clip,keepaspectratio]{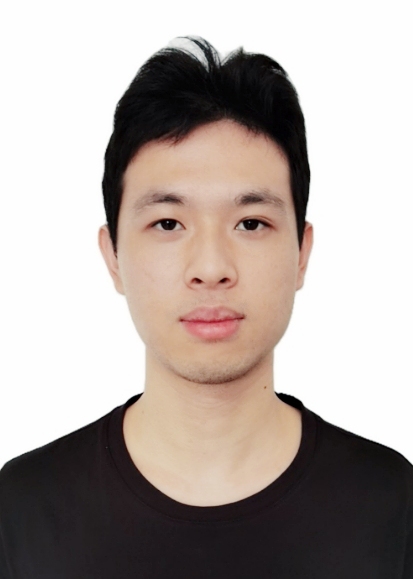}}]
{Shixuan Liu}
received the B.S. degree in systems
engineering from the National University of Defense Technology, Changsha, China, in 2019, where he is currently pursuing the Ph.D degree. His research interests include reinforcement learning, knowledge reasoning and causal discovery.
\end{IEEEbiography}
% \vspace{-33pt}

\begin{IEEEbiography}[{\includegraphics[width=1in,height=1.25in,clip,keepaspectratio]{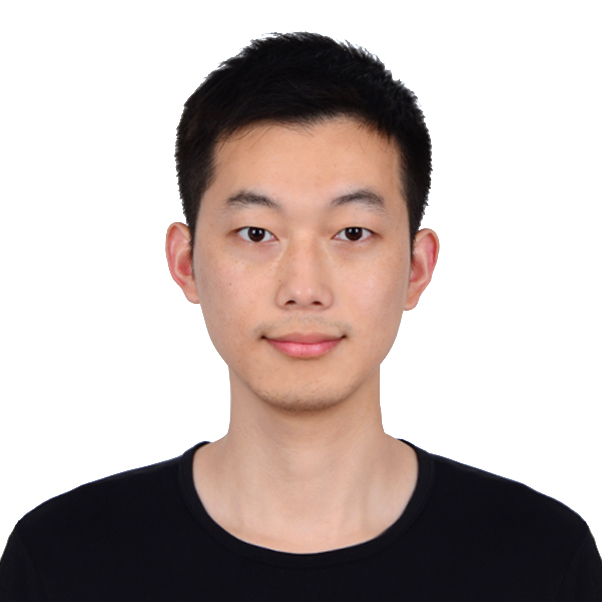}}]{Changjun Fan}
received the B.S. degree, M.S. degree and PhD degree all from National University of Defense Technology, Changsha, China, in 2013, 2015 and 2020. He is also a visiting scholar at Department of Computer Science, University of California, Los Angeles, for two years.  He is currently an associate professor at National University of Defense Technology, China. His research interests include deep graph learning and complex systems, with a special focus on their applications on intelligent decision making. During his previous study, he has published a number of refereed journals and conference proceedings, such as Nature Machine Intelligence, Nature Communications, AAAI, CIKM, etc. 
\end{IEEEbiography}

\begin{IEEEbiography}[{\includegraphics[width=1in,height=1.25in,clip,keepaspectratio]{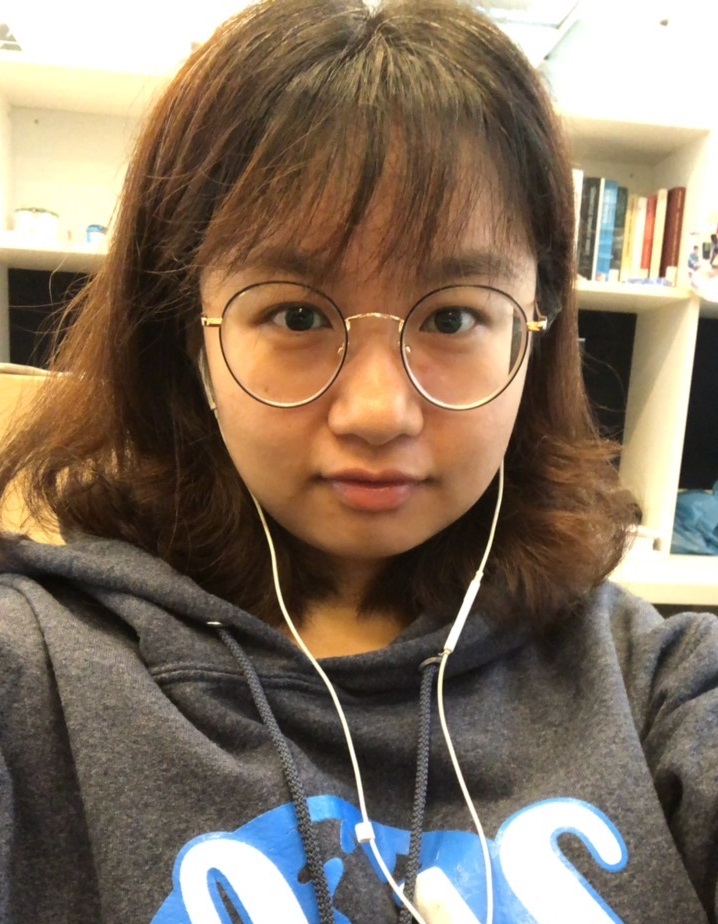}}]{Kewei Cheng}
is a Ph.D. student at UCLA. Prior to joining UCLA, she obtained her master’s degree from Arizona State University and 
her bachelor's degree from Sichuan University. Her research interests are generally in data mining, artificial intelligence, and machine learning, with a focus on neural-symbolic knowledge graph reasoning.
\end{IEEEbiography}
% \vspace{-33pt}

\begin{IEEEbiography}[{\includegraphics[width=1in,height=1.25in,clip,keepaspectratio]{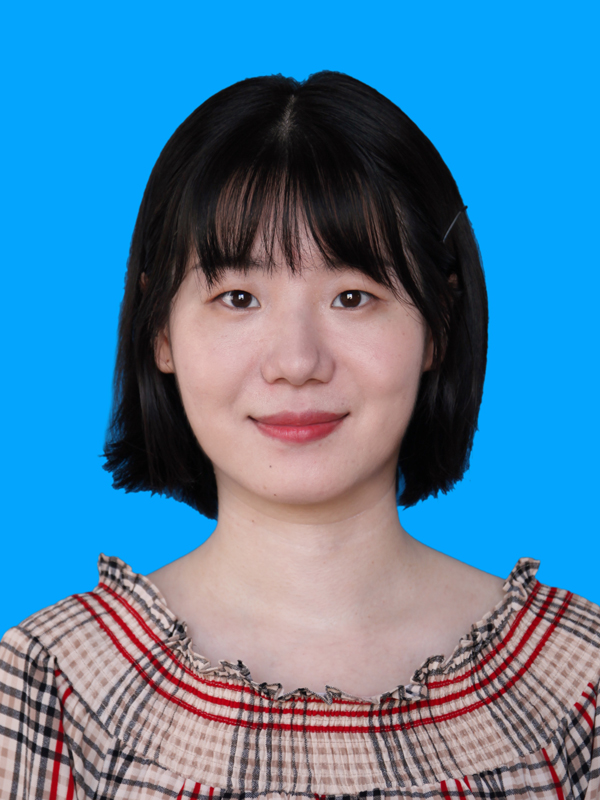}}]{Yunfei Wang}
received the B.S. degree in civil
engineering from the Hunan University, Changsha, China, in 2020. She is now pursuing the Ph.D degree at the National University of Defense Technology, Changsha, China. Her research interests include auto penetration test, reinforcement learning and cyber-security.
\end{IEEEbiography}
% \vspace{-33pt}

% \begin{IEEEbiography}
% [{\includegraphics[width=1in,height=1.25in,clip,keepaspectratio]{./Figures/People/J_Kang.jpg}}]{Jie Kang}
% received the B.S. degree in marketing from Wuhan Textile University, Wuhan, China, in 2020. She is now pursuing the Master's degree in Systems Engineering at National University of Defense Technology. Her research interests include graph adversarial attacks and anomaly detection.
% \end{IEEEbiography}
% % \vspace{-33pt}

% \begin{IEEEbiography}
% [{\includegraphics[width=1in,height=1.25in,clip,keepaspectratio]{./Figures/People/Y_Si.jpg}}]{Yuehang Si}
% received the B.S. degree in systems
% engineering from the National University of Defense Technology, Changsha, China, in 2022, where he is currently pursuing the Ph.D degree. His research interests include deep learning, knowledge reasoning and data analysis.
% \end{IEEEbiography}
% % \vspace{-33pt}

\begin{IEEEbiography}[{\includegraphics[width=1in,height=1.25in,clip,keepaspectratio]{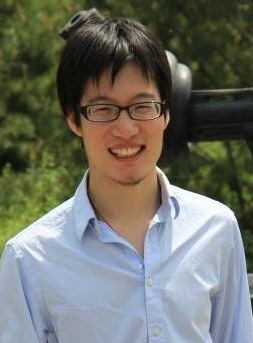}}]{Peng Cui}
(Member, IEEE) is an Associate Professor with tenure in Tsinghua University. He got his PhD degree from Tsinghua University in 2010. His research interests include causally-regularized machine learning, network representation learning, and social dynamics modeling. He has published more than 100 papers in prestigious conferences and journals in data mining and multimedia. Now his research is sponsored by National Science Foundation of China, Samsung, Tencent, etc. He also serves as guest editor, co-chair, PC member, and reviewer of several high-level international conferences, workshops, and journals.
\end{IEEEbiography}

\begin{IEEEbiography}[{\includegraphics[width=1in,height=1.25in,clip,keepaspectratio]{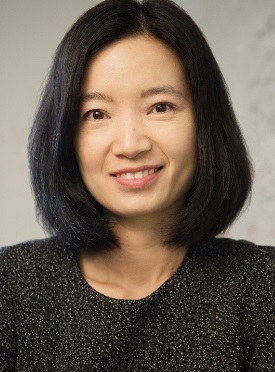}}]{Yizhou Sun}
(Member, IEEE) is an associate professor at Computer Science, UCLA. Prior to that, she joined Northeastern University as an assistant professor in 2013. She received her Ph.D. degree from the Computer Science Department, University of Illinois at Urbana Champaign (UIUC) in December 2012. She got her master's degree and bachelor's degrees in Computer Science and Statistics from Peking University, China.
\end{IEEEbiography}
% \vspace{-33pt}

\begin{IEEEbiography}[{\includegraphics[width=1in,height=1.25in,clip,keepaspectratio]{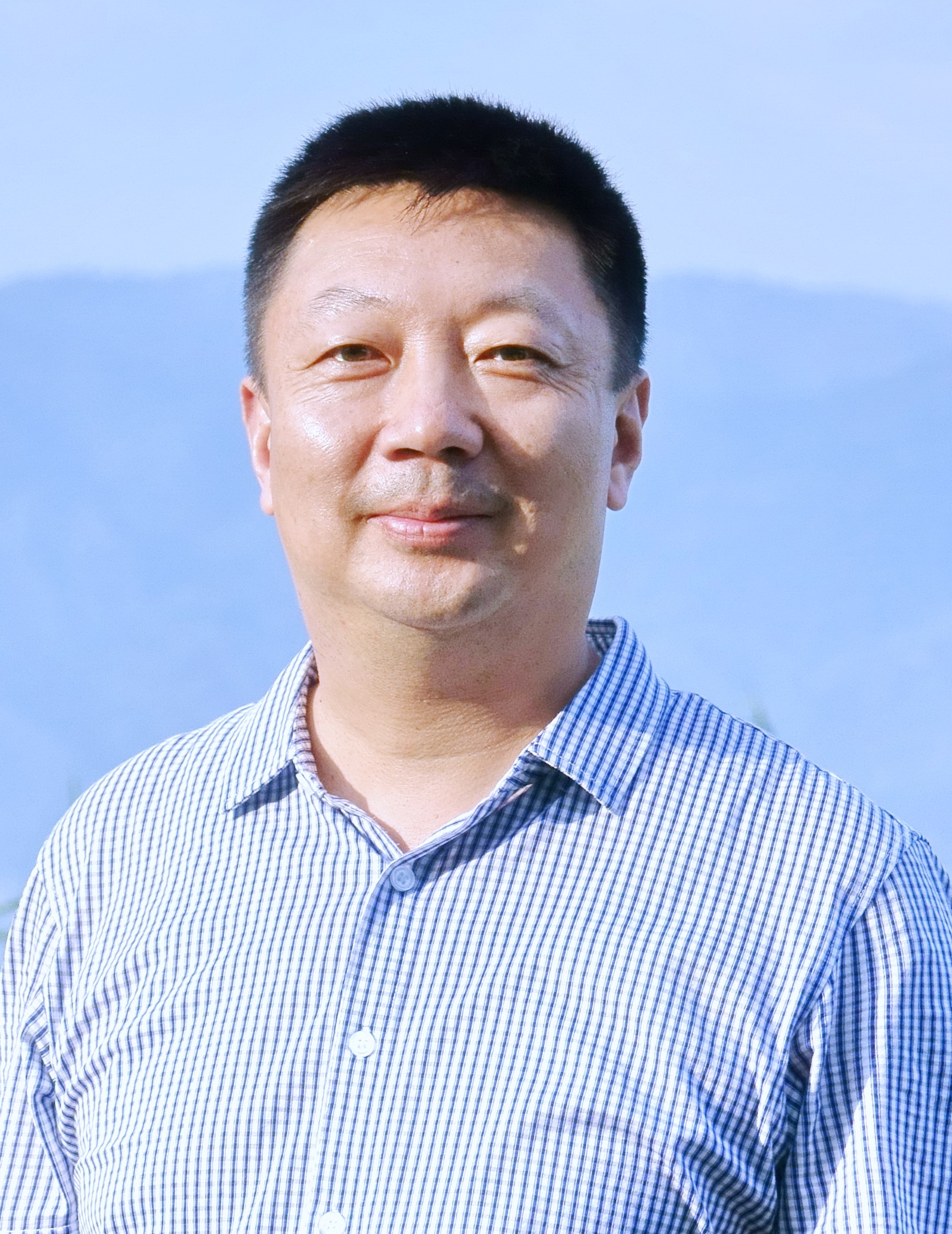}}]{Zhong Liu}
received the B.S. degree in Physics from Central China Normal University, Wuhan, Hubei, China, in 1990, the M.S. degree in computer software and the Ph.D. degree in management science and engineering both from National University of Defense Technology, Changsha, China, in 1997 and 2000. He is a professor in the College of Systems Engineering, National University of Defense Technology, Changsha, China. His research interests include intelligent information systems, and intelligent decision making.
\end{IEEEbiography}

\vfill

% \newpage
\clearpage
\appendices
\section{Baseline Implementations}
\label{appendix_baseline}

\begin{itemize}
\item \textbf{MPDRL~\cite{wan2020mpdrl}.} MPDRL summarizes meta-paths after finding path instances by an RL agent. We use the implementation released by Wan et al.\footnote{github.com/mxz12119/MPDRL} with the reported parameters.

\item \textbf{PCRW~\cite{lao2010pcrw}.} PCRW mines meta-paths based on random walk. We use the implementation released along with \cite{wan2020mpdrl}.

\item \textbf{Autopath~\cite{yang2018autopath}.} Autopath\footnote{github.com/yangji9181/AutoPath} models the similarity between pairs as the empirical probabilities of arrival at the tail entity with the trained model. We fit the train set and the generated samples into the model and tune the parameters to report the best results.

\item  \textbf{Metapath2Vec~\cite{dong2017metapath2vec}.}  Metapath2Vec uses meta-path-based random walks to construct node embeddings\footnote{ericdongyx.github.io/metapath2vec/m2v.html}. 

\item  \textbf{HIN2Vec~\cite{fu2017hin2vec}.} We use the code from the authors\footnote{github.com/csiesheep/hin2vec} with our highly-tuned hyperparameters.

\item \textbf{NSHE~\cite{zhao2020network}.} NSHE is a network schema preserving HIN embedding method that retains high-order structure.\footnote{github.com/Andy-Border/NSHE}

\item \textbf{RotatE~\cite{sun2019rotate}.} RotatE\footnote{github.com/DeepGraphLearning/KnowledgeGraphEmbedding} learns embeddings to represent entities and relations in KBs. We adopt the reported hyperparameters for Yago and apply the same for other datasets.

\item \textbf{TransE~\cite{bordes2013transe}.} TransE builds the embeddings of instance triples by making the sum of the head/relation vector close to the tail vector. We use an implementation by Han et al.\footnote{github.com/thunlp/OpenKE}.

\item \textbf{DistMult~\cite{yang2014Distmult}.} DistMult associates related entities using Hadamard product of embeddings$^5$. 

\item \textbf{ComplEx~\cite{trouillon2016complex}.}
ComplEx migrates DistMult in a complex space and offers comparable performance\footnote{github.com/ttrouill/complex}.

\item \textbf{RNNLogic~\cite{qu2020rnnlogic}.}
RNNLogic is a rule-learning method that consists of a recurrent neural network to encode KB into a low-dimensional representation and a logic reasoning module to learn and infer logic rules from the encoded representation\footnote{github.com/DeepGraphLearning/RNNLogic}.

\item  \textbf{MINERVA~\cite{das2018minerva}.} MINERVA is a neural RL-based multi-hop approach for automated reasoning\footnote{github.com/shehzaadzd/MINERVA}. For per-relation experiments, similar to their evaluation approach on NELL that uses the predicted scores (logits) for head/tail entity pairs, we calculate the predicted scores for all positive and negative samples and obtain the similarity with a softmax operation over these scores, which produces better result than using the original scores.

\item \textbf{MLN4KB~\cite{fang2023mln4kb}.} MLN4KB is a method based on Markov logic network that can leverage the sparsity of knowledge bases\footnote{github.com/baidu-research/MLN4KB}. 
\end{itemize}

\section{Dataset Details}
\begin{table}[h!]
    \centering
       \caption{Statistics of real-world datasets}
\resizebox{\linewidth}{!}{%
 \begin{tabular}{c|ccc|c}
    \toprule[1.5pt]
        \multirow{2}{*}{Dataset} &\multicolumn{3}{c|}{Instance Graph} &Schema Graph
       \\
    &\#Entity&\#Relation &\#Triples&\#Entity Types\\
     \hline
        Yago26K-906 & 26,078 & 34 & 390,738 & 906 \\
        DB111K-174&  111,762 & 305 & 863,643&174\\
      	NELL&  49,869 & 827 & 296,013&756\\
      	Chem2Bio2RDF&  295,911 & 11 & 727,997&9\\
       \bottomrule[1.5pt]
   \end{tabular}
   }
   \label{table_dataset}
\end{table}
\label{appendix_dataset}
\begin{itemize}
\item \textbf{YAGO26K-906~\cite{suchanek2007yago,hao2019joie}.} Yago is a KB built on extracted facts from Wikipedia and WordNet. The original Yago poses a limitation on the semantic relations among entity types. We adopt the preprocessed core Yago facts with enriched taxonomy, released by Hao et al.\footnote{github.com/JunhengH/joie-kdd19}.

\item \textbf{DB111K-174~\cite{auer2007dbpedia}.} The original Dbpedia is a large KB extracted from Wikipedia involving considerable specific domains and general knowledge. Similar with YAGO26K-906, we use the taxonomy-enriched version$^{10}$.

\item \textbf{NELL~\cite{mitchell2018never}.} NELL is a KB constructed from 500 million unstructured web pages. We utilize the preprocessed 1115-th portion of NELL, available in ~\cite{wan2020mpdrl}. 

\item \textbf{Chem2Bio2RDF~\cite{chen2010chem2bio2rdf}.} Chem2Bio2RDF is a schema-simple HIN linking drug
candidates and their biological annotations for drug-target prediction\footnote{chem2bio2rdf.org/}.
\end{itemize}

The statistics of the datasets are summarized in Table \ref{table_dataset}.

\begin{table*}[htb]
\centering
\caption{Hyper-parameters}\label{table_hyperparamenters}
\resizebox{0.7\linewidth}{!}{%
\begin{tabular}{c|c|c|m{5cm}}
\toprule[1.5pt]
\textbf{Hyper-parameter}& \textbf{Symbol}& \textbf{Value} & \multicolumn{1}{c}{\textbf{Description}}
\\ \midrule[1.2pt]
\multirow{4}{*}{Base Iterations} &\multirow{4}{*}{$I_{base}$}                              & 500 (YAGO26K-906)       & \multirow{4}{*}{Base iterations for training}                   \\
&&500 (DB111K-906)&\\
&&1000 (NELL)&\\ 
&&50 (Chem2Bio2RDF)&\\
\hline
Rounds &$I_{r}$                             & 5       & Training rounds for each relation     
\\
\hline
Meta-path Length &$l$                              & 5            & Maximum length for meta-paths       \\
\hline
\multirow{4}{*}{Hidden Size}   &\multirow{4}{*}{$d_h$}                              & 200 (YAGO26K-906)           & \multirow{4}{*}{Dimension of hidden layers output}                                                     \\
&&400 (DB111K-906)&
\\
&&200 (NELL)&
\\
&&100 (Chem2Bio2RDF)\\
\hline
Embedding Size     &$d_e$        & 64             & Dimension of embeddings for concepts and relations representation                                                                                                  \\
\hline
\multirow{4}{*}{Batch Size}    &\multirow{4}{*}{$K$}                       & 20 /80 (YAGO26K-906)         & \multirow{4}{5cm}{Batch size to sample concept pairs to form queries (batch size for multi-relation / per-relation if two values are given)}
\\
&&20 (DB111K-906)& 
\\
&&40 (NELL)& 
\\
&&80 (Chem2Bio2RDF)& 
\\
\hline
Rollouts     &$N$                         & 40             & Number of rollouts for each training sample    
\\
\hline
\multirow{4}{*}{Test Batch Size}     &\multirow{4}{*}{}                         & 10 (YAGO26K-906)             & \multirow{4}{5cm}{Batch size during test}    
\\
&&40 (DB111K-906)& 
\\
&&40 (NELL)&
\\
&&10 (Chem2Bio2RDF)&
\\
\hline
\multirow{4}{*}{Test Rollouts}     &\multirow{4}{*}{}                         & 10 (YAGO26K-906)             & \multirow{4}{5cm}{Beam Search Width}    
\\
&&40 (DB111K-906)& 
\\
&&40 (NELL)&
\\
&&10 (Chem2Bio2RDF)&
\\
\hline
Learning Rate &$\alpha$& 0.0005& Update rate for Adam optimizer
\\
\hline
Beta &$\beta$&0.05 (decay rate 0.9) & Coefficient for entropy regularization term
\\
\hline
\multirow{2}{*}{Lambdas} & \multirow{2}{*}{$\lambda_1, \lambda_2$} &10, 1 (Multi-Relation) & \multirow{2}{5cm}{Weight in Eq.\ref{equ_reward}} \\
&&5, 1 (Per-Relation)& 
\\ \bottomrule[1.5pt]
\end{tabular}}%
\end{table*}

\noindent \textbf {Train/Test Relations for YAGO26K-906:}
\begin{itemize}
    \item \textbf{Train}: isLocatedIn, influences, isAffiliatedTo, diedIn, hasWonPrize, hasCapital, happenedIn, owns, graduatedFrom, isMarriedTo, isConnectedTo, isPoliticianOf, participatedIn, created, actedIn, worksAt, dealsWith, directed, wroteMusicFor, hasAcademicAdvisor, hasNeighbor, isKnownFor, edited, isInterestedIn.
    \item  \textbf{Test}: wasBornIn, isCitizenOf, isLeaderOf, hasChild, playsFor, livesIn.
\end{itemize}

\noindent \textbf {Train/Test Relations for DB111K-174:}
\begin{itemize}
    \item \textbf{Train}: previousEvent, predecessor, owningCompany, parentCompany, owner, formerBandMember, sire, child, successor, division, designCompany, deputy, associatedMusicalArtist, riverMouth, executiveProducer, nextEvent, related, position, designer, foundationPlace, artist, creator, citizenship, doctoralStudent, mouthPlace, associatedBand, governor, associate, doctoralAdvisor, sourcePlace, sisterStation, largestCity, sourceMountain, league, subsidiary, previousWork, developer, mouthMountain, lieutenant, bandMember, relative.
    \item  \textbf{Test}: musicalBand, musicalArtist, subsequentWork, nationality, spouse, countySeat, stateOfOrigin, distributingCompany, distributingLabel, parent, trainer.
\end{itemize}

\section{Supporting Node Classification}
\begin{table}[htb]
\label{tab:node_classification}
\caption{Node classification task on ACM and DBLP.}
\resizebox{\linewidth}{!}{%
\begin{tabular}{ccccc}
\toprule[1.5pt]
 & \multicolumn{2}{c}{ACM} & \multicolumn{2}{c}{DBLP} \\
 & Marco-F1   & Mirco-F1   & Marco-F1    & Mirco-F1   \\ \hline
DeepWalk~\cite{perozzi2014deepwalk}         & 84.17      & 83.99      & 84.73       & 85.22      \\
Metapath2Vec~\cite{dong2017metapath2vec}     & 73.83      & 73.67      & 91.92       & 92.73      \\
RGCN~\cite{schlichtkrull2018rgcn}             & 91.56      & 91.37      & 91.58       & 92.03      \\
GAT~\cite{velivckovic2017graph}              & 87.57      & 87.38      & 90.79       & 91.85      \\
HAN~\cite{wang2019heterogeneous}              & 90.93      & 90.78      & 91.56       & 91.96      \\
MAGNN~\cite{fu2020magnn}            & 90.83      & 90.75      & 92.93       & 93.51      \\
HGT~\cite{hu2020heterogeneous}              & 91.18      & 90.98      & 92.89       & 93.46      \\ \hline
SchemaWalk-HAN   & 92.48      & 92.34      & 93.26       & 93.68      \\
SchemaWalk-MAGNN & \textbf{92.96}      & \textbf{92.82}       & \textbf{93.37}       & \textbf{93.83}     
\\ \bottomrule[1.5pt]
\end{tabular}}
\end{table}
\noindent While our primary focus is on relation prediction, our method could be combined with existing HIN embedding methods for performing node classification. Specifically, we employ~\model~to identify long meta-path, which are subsequently used to embed HIN with HAN~\cite{wang2019heterogeneous} and MAGNN~\cite{fu2020magnn}. 
We follow the basic node classification task as specified in~\cite{wang2019heterogeneous} and we use 30\% of the target type nodes for training and 70\% for testing. As evidenced in Table~\ref{tab:node_classification}, our method effectively identifies lengthy meta-paths, which when used in conjunction with HAN and MAGNN, yield favourable performance for node classification.

\section{Further Training Settings} \label{appendix_hyper}
We conduct experiments on a desktop computer with a 10-core CPU, a 32GB memory and a 12GB RTX-2080Ti GPU. The schema-level embeddings are randomly initialized. We use sparse matrix to compute coverage and confidence for all necessary meta-paths during training. The calculated numerical values for the explored meta-paths are stored in memory during training to reduce unnecessary computation. 
% \textcolor{red}{The memory consumption varies in association with the size of the schema graph, typically ranging between 2GB and 4GB.}
See Table~\ref{table_hyperparamenters} for hyper-parameters.

\section{Complexity Analysis} 
\label{appendix_complex}

\subsection{Train Time Complexity}
\label{appendix_time_train_complex}
The time complexity of our training process stems from the following two components: LSTM-based policy network training and the computation of coverage and confidence.
\begin{itemize}
    \item Assuming we train with an overall batch size of $B=KN$ (concept pair batch size multiplied by number of rollouts), and each meta-path possesses a maximum length $l+1$. We can discern the time complexity for LSTM training per epoch as $\mathcal{O}(Bl)$~\cite{zhou2021informer}. 
    \item For the latter, the complexity is largely rooted in the multiplication of sparse adjacency matrices, employed to derive the quantity of nodes pairs connected via a meta-path. 
    This multiplication process for two sparse matrices, denoted as $A_{n\times m}$ and $B_{n'\times m'}$, is carried out utilizing the sparse matrix-matrix multiplication algorithm~\cite{bank1993sparse}, implemented in the Scipy package. The resulting complexity for multiplication of two sparse matrices is presented as $\mathcal{O}(n*S^2 + max(n,m'))$, where $S$ is the maximum number of nodes in a row of A or column of B.
    In our case, $n=m'=|V|$, where $|V|$ is the number of nodes in the instance graph. Besides, $S$ can be bounded by the maximum degree $\Delta\mathcal{G}_I$ of the instance graph $\mathcal{G}_I$. Combining these facts, we get the complexity to evaluate a single meta-path is capped by $\mathcal{O}((|V|(1+\Delta(\mathcal{G}_I)^2))^{l-1})\approx \mathcal{O}((|V|\Delta(\mathcal{G}_I)^2)^{l-1})$. 
    During the training process, it is possible that duplicate, invalid or previously evaluated meta-paths may emerge, which would not necessitate further re-evaluation. Notwithstanding, in a worst-case scenario, all generated meta-paths within one epoch could require evaluation.
\end{itemize}

% To multiply two sparse matrices $A_{n\times m}$ and $B_{n'\times m'}$, we use the Scipy package based on the the SMMP algorithm~\cite{bank1993sparse}, which renders the complexity $\mathcal{O}(n*K^2 + max(n,m'))$ where $K$ is the maximum number of nodes in a row of A and column of B.
Considering all these factors, we can conclude that the overall complexity for the training process is bounded by $\mathcal{O}(B(l+(|V|\Delta(\mathcal{G}_I)^2)^{l-1})$ per epoch.
% The adjacency matrix is denoted as $R(T_1, r_1, T_2) \in \{0,1\}^{|V|\times |V|}$ whose element $R_{i,j}$ denotes whether an edge exists between node $i$ and node $j$. 

\subsection{Inference Time Complexity} 
\label{appendix_infer_time_complexity}
With reference to~\cite{das2018minerva}, assuming a power law degree distribution for the schema graph $\mathcal{G}_s$, a pattern seen frequently in natural graphs, we can deduce that the average inference time for a single query per step with our method approximates $\mathcal{O}(\frac{\eta}{\eta-1})$, provided that the coefficient of the power law $\eta>1$. The median inference time remains $\mathcal{O}(1)$ for all values of $\eta$. Therefore, assuming we infer meta-paths of length $l+1$ with a batch size of $B'=K'N'$, the complexity of the test time is expressed as $\mathcal{O}(B'l \frac{\eta}{\eta-1})$ per epoch.

\subsection{Space Complexity} 
During training, the space complexity is mainly based on the storage requirements of the input tensors, the LSTM-based policy network and the sparse matrix multiplication (SMM) operations. Suppose the overall batch size is $B=KN$ (concept pair batch size multiplied by number of rollouts) and the meta-path is of maximum length $l+1$. $d_e$ and $d_h$ are respectively the dimension size of embeddings and hidden layers.

\begin{itemize}
    \item The input tensors contains schema-level representations expressed in Equation (\ref{equ_enc}), including $[{\bf r}_{i-1}\mathbin\Vert {\bf t}_i]$, and ${\bf enc}_i=[{\bf S}_i\mathbin\Vert {\bf t}_i \mathbin\Vert {\bf r}_q \mathbin\Vert ({\bf t}_{\rm tgt}-{\bf r}_q)]$, and the feature matrix for decision $D_i$ specified in Equation (\ref{equ_dec}). The items in Equation (\ref{equ_enc}) result in space complexity of $\mathcal{O}(6Bld_e)$ and the decision feature matrix renders space complexity of $\mathcal{O}(Bl\langle \mathcal{G}_S \rangle d_e)$, where $\langle \mathcal{G}_S \rangle$ is the mean degree of the schema graph $\mathcal{G}_S$. The temporary history vector consumes $\mathcal{O}(Bd_h)$ space. Altogether, this part produces space complexity of $\mathcal{O}(B(l(6+\langle \mathcal{G}_S \rangle )d_e+d_h))$.
    
    \item For an LSTM, the model parameters include four weight matrices and four bias vectors for the forget, input, output and cell states. The space complexity for the parameters alone is $\mathcal{O}(4d_ed_h+4{d_h}^2)$. For many optimization algorithms such as Adam, we also generally need to store additional per-parameter state, which would be the same order, leading to overall $\mathcal{O}(8d_ed_h+8{d_h}^2)$ space complexity for the parameters. In backward training, we need to store the cell states and hidden states at each time step for every sequence in a batch, and the space complexity for these variables is $\mathcal{O}(Bld_h)$. Altogether, the LSTM part produces space complexity of $\mathcal{O}(8d_ed_h+8{d_h}^2+Bld_h)$. 
    For the linear decoder which produces probability to select an outgoing edges among candidate edges, the output dimension is bounded by $\Delta(\mathcal{G}_S)$, and therefore the space complexity for the parameters alone is $\mathcal{O}(d_h^2+d_h\Delta(\mathcal{G}_S))$. As Adam store additional per-parameter state, leading to a doubling of the parameter memory requirement, which gives $\mathcal{O}(2d_h^2+2d_h\Delta(\mathcal{G}_S))$. During forward propagation, we need to store the inputs and outputs of each layer for use in back-propagation, leading to approximately $\mathcal{O}(N(2d_h+\Delta(\mathcal{G}_S)))$. Altogether, the linear part produces space complexity of $\mathcal{O}(2d_h^2+2d_h\Delta(\mathcal{G}_S)+N(2d_h+\Delta(\mathcal{G}_S)))$.
    % The network parameter accounts for a space complexity that is $\mathcal{O}(l)$. 
\end{itemize}
The aforementioned components occupy GPU memory that varies from 700MB to 2.4GB, dependent on the dataset.

\begin{itemize}
    \item For the SMM part, storing sparse matrices for a meta-path requires $\mathcal{O}(\sum_i^L{nnz_i})\approx \mathcal{O}(l*\langle \mathcal{G}_I \rangle)$ storage, where $nnz_i$ is the number of non-zero items of a matrix $i$. Additionally, the SMM algorithm used in our method requires temporary storage on the magnitude of $\mathcal{O}(|V|)$. The overall space complexity for SMM is $\mathcal{O}(|V|+l\langle \mathcal{G}_I \rangle)$
\end{itemize}

The above part incurs a CPU memory usage typically between 2GB and 4GB, dependent on the dataset. During the inference stage, the space complexity related to SMM is absent as it is not required.

\end{document}